\documentclass[10pt,twocolumn,letterpaper]{article}

\usepackage[pagenumbers]{cvpr} % To force page numbers, e.g. for an arXiv version

\usepackage[utf8]{inputenc}
\usepackage[T1]{fontenc}

\definecolor{cvprblue}{rgb}{0.21,0.49,0.74}
\usepackage[pagebackref,breaklinks,colorlinks,allcolors=cvprblue]{hyperref}

\def\paperID{*****} % *** Enter the Paper ID here
\def\confName{CVPR}
\def\confYear{2025}

\title{\name{}: Bridging Cinematic Principles and Generative AI for Automated Film Generation}

\author{Kaiyi Huang$^{1*}$
\qquad
Yukun Huang$^{1}$
\qquad 
Xintao Wang$^{2\dagger}$
\qquad
Zinan Lin$^{3}$
\qquad
Xuefei Ning$^{4}$
\\
\qquad
Pengfei Wan$^{2}$
\qquad
Di Zhang$^{2}$
\qquad
Yu Wang$^{4}$
\qquad
Xihui Liu$^{1\dagger}$
\vspace{0.2cm}
\\
{\normalsize $^{1}$ The University of Hong Kong \quad $^{2}$ Kuaishou Technology}
{\normalsize \quad $^{3}$ Microsoft Research} \quad \normalsize$^{4}$ Tsinghua University \\
\href{https://filmaster-ai.github.io}{https://filmaster-ai.github.io}}

\usepackage{booktabs} % For formal tables

\usepackage[ruled]{algorithm2e} % For algorithms
\usepackage{graphicx}

\usepackage{array} % Explicitly load array for more control, arydshln often needs it
\usepackage{pifont}   % For \ding{51} (checkmark) and \ding{55} (cross) - alternative

\usepackage{makecell} % For better cell formatting (e.g., line breaks)
\usepackage{graphicx} % For \resizebox
\usepackage{multirow} % For multirow cells
\usepackage{colortbl}
\usepackage{placeins}

\usepackage{tcolorbox}
\definecolor{chart}{HTML}{1f77b4}
\newtcolorbox{example}[1][]{
  colback=chart!5!white,
  colframe=chart,
  floatplacement=tp*,  % 修改这里！使用 tp* 或 Htp* 或其他带星号的浮动参数
  title=\centering\bfseries\sffamily #1, % 稍微调整了标题格式，使其更像标题
  fonttitle=\bfseries\sffamily, % 确保标题字体
  width=\textwidth, % 通常浮动体*会自动处理，但有时显式指定有帮助
  nobeforeafter, % 尝试添加这个
}

\definecolor{amethyst}{rgb}{0.6, 0.4, 0.8}

\definecolor{lemon}{RGB}{255,247,0}
\definecolor{maize}{RGB}{250,237,94}
\definecolor{coco1}{HTML}{D9E4EC}
\definecolor{coco2}{HTML}{B7CFDC}
\definecolor{coco3}{HTML}{6AABD2}
\definecolor{coco4}{HTML}{385E72}
\hypersetup{
    colorlinks=true,
    linkcolor=coco3,
    filecolor=coco4,      
    urlcolor=coco3,
    citecolor=coco3,
}

\definecolor{mustard}{RGB}{255,219,89}
\definecolor{ocre}{RGB}{241,103,35}
\definecolor{Tangerine}{RGB}{253,128,8}
\definecolor{framegreen}{RGB}{153, 188, 133}
\definecolor{bggreen}{RGB}{235, 250, 228}
\definecolor{lightgray}{RGB}{239,240,241}
\definecolor{c0}{cmyk}{1,0.3968,0,0.2588} 
\definecolor{c1}{cmyk}{0,0.6175,0.8848,0.1490} 
\definecolor{c2}{cmyk}{0.1127,0.6690,0,0.4431} 
\definecolor{c3}{cmyk}{0.3081,0,0.7209,0.3255} 
\definecolor{c4}{RGB}{164, 16, 52}

\definecolor{g0}{HTML}{ffdbdc}
\definecolor{g1}{HTML}{ffefeb}
\definecolor{g2}{HTML}{FCFFEB}
\definecolor{g3}{HTML}{e9fcea}
\definecolor{g4}{HTML}{d4f8d4}
\definecolor{c0}{cmyk}{1,0.3968,0,0.2588} 
\definecolor{c1}{cmyk}{0,0.6175,0.8848,0.1490} 
\definecolor{c2}{cmyk}{0.1127,0.6690,0,0.4431} 
\definecolor{c3}{cmyk}{0.3081,0,0.7209,0.3255} 
\definecolor{c4}{RGB}{164, 16, 52}
\definecolor{orange}{HTML}{E66100}
\definecolor{bluex}{HTML}{0C7BDC}
\definecolor{yellow}{HTML}{FFC20A}
\definecolor{lightpurple}{HTML}{E6E6FA}
\definecolor{lightbluee}{HTML}{e8f4f8}
\definecolor{c5}{HTML}{EE4E4E}
\definecolor{gggggg}{HTML}{EFEFEF}
\newtcbox{\hlprimarytab}{on line, box align=base, colback=blue!12,colframe=white,size=fbox,arc=3pt, before upper=\strut, top=-2pt, bottom=-4pt, left=-2pt, right=-2pt, boxrule=0pt}
\newtcbox{\hlsecondarytab}{on line, box align=base, colback=orange!10,colframe=orange,size=fbox,arc=3pt, before upper=\strut, top=-2pt, bottom=-4pt, left=-2pt, right=-2pt, boxrule=0pt}
\newtcbox{\hlwhite}{on line, box align=base, colback=blue!12,colframe=white,size=fbox,arc=2pt, before upper=\strut, top=-2pt, bottom=-4pt, left=-2pt, right=-2pt, boxrule=0pt}
\newtcbox{\hlyellow}{on line, box align=base, colback=BlueGreen!10,colframe=white,size=fbox,arc=2pt, before upper=\strut, top=-2pt, bottom=-4pt, left=-2pt, right=-2pt, boxrule=0pt}

\definecolor{darkercheck}{RGB}{0, 128, 0}     % A darker green for checkmarks
\definecolor{darkercross}{RGB}{192, 0, 0}     % A darker red for crosses
\definecolor{mycolor_green}{HTML}{E6F8E0}
\definecolor{mycolor_blue}{HTML}{E7EFFA}
\definecolor{mycolor_orange}{HTML}{EEA851}
\definecolor{mycolor_deepblue}{HTML}{4396A4}
\definecolor{mycolor_gray}{HTML}{DAD7D7}

\newcommand{\cmark}{\textcolor{darkercheck}{\checkmark}}
\newcommand{\xmark}{\textcolor{darkercross}{$\times$}}

\newcommand{\name}{FilMaster}
\newcommand{\eval}{FilmEval}
\newcommand{\generation}{Reference-Guided Generation Stage}
\newcommand{\coordination}{Generative Post-Production Stage}
\newcommand{\camera}{Multi-shot Synergized RAG Camera Language Design}
\newcommand{\rhythm}{Audience-Centric Cinematic Rhythm Control}
\begin{document}

\twocolumn[{
\renewcommand\twocolumn[1][]{#1}
\maketitle

\begin{center} 
\vspace{-2em}
    % \fbox{\rule{0pt}{3in} \rule{.9\linewidth}{0pt}}
    \includegraphics[width=\linewidth]{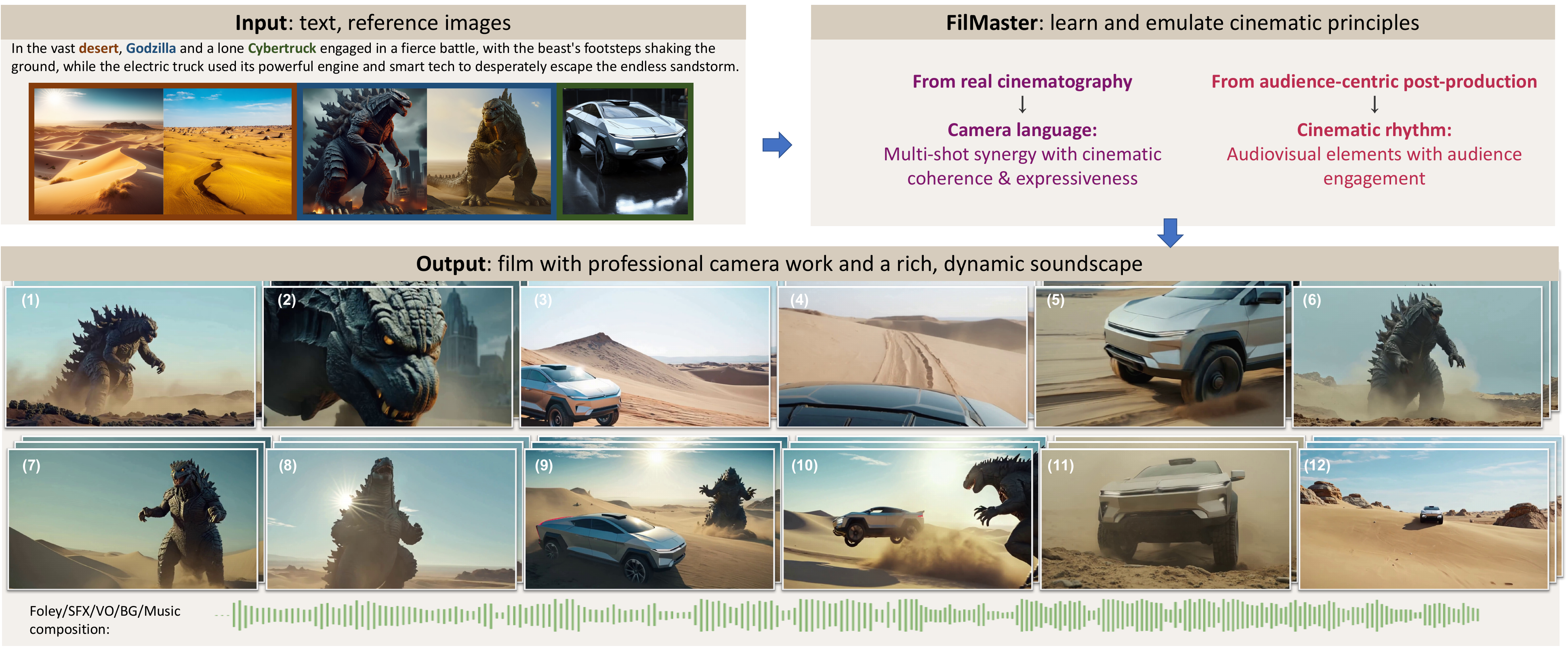}
    \captionsetup{type=figure}
    \vspace{-1em}
      \caption{Video samples generated by \name{}. Using a textual theme and reference images for characters and locations, \name{} crafts high-quality films complete with professional camera language and cinematic rhythm, including rich, multi-layered audiovisual outputs (foley, sound effects (SFX), voice-over(VO), background ambiance, musical scoring, and video).}

    \label{fig:teaser}
    % \vspace{-1pt}
\end{center}
}]

\let\thefootnote\relax\footnotemark\footnotetext{$^{*}$Part of the work done during intenrship at KwaiVGI, Kuaishou Technology. $^{\dagger}$ Corresponding author.}

\begin{abstract}
AI-driven content creation has shown potential in revolutionizing film production. However, existing film generation systems struggle to understand and implement fundamental cinematic principles and thus fail to generate professional-quality films, particularly lacking diverse and expressive camera language and cinematic rhythm. This often results in templated visuals and unengaging narratives. To address these limitations, we introduce \name{}, a comprehensive end-to-end AI-powered automated film generation system to integrate real-world cinematic principles for professional-grade film generation, yielding editable and industry-standard outputs. \name{} is built upon two key cinematic principles: (1) learning cinematography from extensive real-world film data and (2) emulating professional, audience-centric post-production workflows.
Inspired by these principles, \name{} incorporates two stages: \generation{} which transforms user input to video clips, and \coordination{} which transforms raw footage into audiovisual outputs by orchestrating both visual and auditory elements for cinematic rhythm.
%To realize these principles, \name{} incorporates two key innovations. 
Our \generation{} highlights a \camera{} module to guide the AI system to generate professional and expressive camera language in videos by retrieving reference clips from a vast corpus of 440,000 film clips. % to generate expressive, context-aware camera work with cinematic coherence, overcoming templated approaches. 
Our \coordination{} emulates professional post-production by designing an \rhythm{} module, including a \textit{Rough Cut} assembly, a \textit{Fine Cut} process informed by simulated audience feedback, for the effective integration of audiovisual elements through video editing and sound design,
% and sophisticated multi-scale audio-visual synchronization
to achieve engaging content and emotional impact. The whole system is empowered by generative AI models such as (M)LLMs, and video generation models. Furthermore, we introduce \eval{}, a comprehensive benchmark for evaluating AI-generated films across key cinematic dimensions. Extensive experiments demonstrate \name{}'s superior performance, particularly in sophisticated camera language design and nuanced cinematic rhythm control, paving the way for generative AI in professional filmmaking.

% Despite the immense potential of AI-driven film production, a significant gap remains: current systems often struggle to imbue content with genuine cinematic artistry, particularly in nuanced camera language design and sophisticated cinematic rhythm control, hindering real-world adoption. These limitations primarily stem from an inability to learn from and apply established cinematic principles derived from real-world filmmaking. To address these critical gaps, we introduce \name{}, an end-to-end autonomous system for professional-grade cinematic film creation. \name{} pioneers a novel approach by: (1) employing Retrieval-Augmented Generation (RAG) to learn scene-level, multi-shot camera language from a vast corpus of 440,000 real film clips, ensuring expressive and coherent visual storytelling; and (2) implementing audience-centric, multimodal cinematic rhythm control that emulates professional post-production workflows, including sophisticated editing and rich audio-visual synchronization for compelling narrative and emotional pacing. Furthermore, \name{} generates fully editable, multi-track timelines using the industry-standard OpenTimelineIO (OTIO) format, bridging the gap between AI generation and professional production pipelines. We also establish a comprehensive benchmark, \eval{}, for evaluating AI-generated film quality. Experiments demonstrate \name{}'s superior capabilities in producing high-quality, cohesive cinematic content with sophisticated camera work and engaging rhythm, significantly advancing the state-of-the-art in automated film production.
\end{abstract}

\section{Introduction}
% \TODO{P2: narrative pacing: optimization
% --> change words
% control NO emotional resonance, and audience en-gagement: effects
% in order to audience engagement
% audience engagement sound design + video editing
% 
% unify audience-centric 
% unify narrative flow --> narrative pacing
% unify emotional impact -> emotional resonance
% }

% P1 problems: lack cinematic principles
Traditional film production involves a complex process from pre-production to post-production~\cite{honthaner2013completeFilmProduction}.
The rapid development of generative AI~\cite{kong2024hunyuanvideo, yang2024cogvideox} and (Multimodal) Large Language Models ((M)LLMs)~\cite{gpt-4o, team2023gemini} has revolutionized AI-driven content creation, opening up new possibilities for film production~\cite{zhang2025generativeAIforFilm}. 
However, current AI-driven film generation systems oversimplify the intricate process of filmmaking, often neglecting core cinematic principles~\cite{bordwell2004filmart}. They typically use LLMs for scriptwriting, and then generate sequential video clips independently through visual generation models~\cite{wu2025movieagent, li2024animdirector}. The resulting outputs tend to be templated and unengaging, falling short of professional cinematic standards, which limits their integration into real-world production workflows.
These limitations largely arise because current AI systems often struggle to \textbf{understand and implement fundamental cinematic principles}, when applied to the complex art of filmmaking~\cite{bordwell2004filmart}. These principles are crucial for crafting effective cinematic narratives~\cite{rabiger2013directing}, particularly evident in two key aspects:
(1) Camera Language: the artful use of cinematographic techniques to convey narrative and emotion by crafting the visual language~\cite{mascelli1965fiveCofCinematography, arijon1976grammar}. A lack of effective camera language leads to visuals that are either templated or overly generic, lack of coherence, failing to convey the intended narrative. For instance, while MovieAgent~\cite{wu2025movieagent} attempts to integrate camera language, it often produces incoherent or uninspired shots, as these elements are generated based on LLM's imagination rather than professional cinematic techniques, resulting in incoherence and a lack of expressiveness. 
(2) Cinematic Rhythm: the masterful orchestration of pacing and flow through editing and sound to shape the audience's emotional journey~\cite{ruszev2018rhythmic, murch2001blink}. The absence of such control results in a flat, unengaging experience, where video clips are often simply concatenated with limited and desynchronized audio~\cite{wu2025movieagent, LTXStudioAppMisc} (\Cref{tab:intro_compare}). This leads to repetitive narration and a diminished emotional impact for the audience.
Consequently, without a deep comprehension of how cinematic techniques contribute to films and how rhythm shapes audience experience, AI-generated content frequently yields uninspired visuals and fails to capture the deliberate artistic choices of skilled filmmakers.

% However, integrating cinematic principles in professional film production presents profound challenges: \textit{How to generate scene-coherent, expressive camera language?} \textit{How to master the art of cinematic rhythm, and make effective, emotionally resonant adjustments?}

\begin{figure}[t]   % htbp
  \centering
    % \fbox{\rule{0pt}{1.5in} \rule{1\linewidth}{0pt}}
   % \includegraphics[width=\linewidth]{figures/multi_agent_v5.pdf}
   \includegraphics[width=\linewidth]{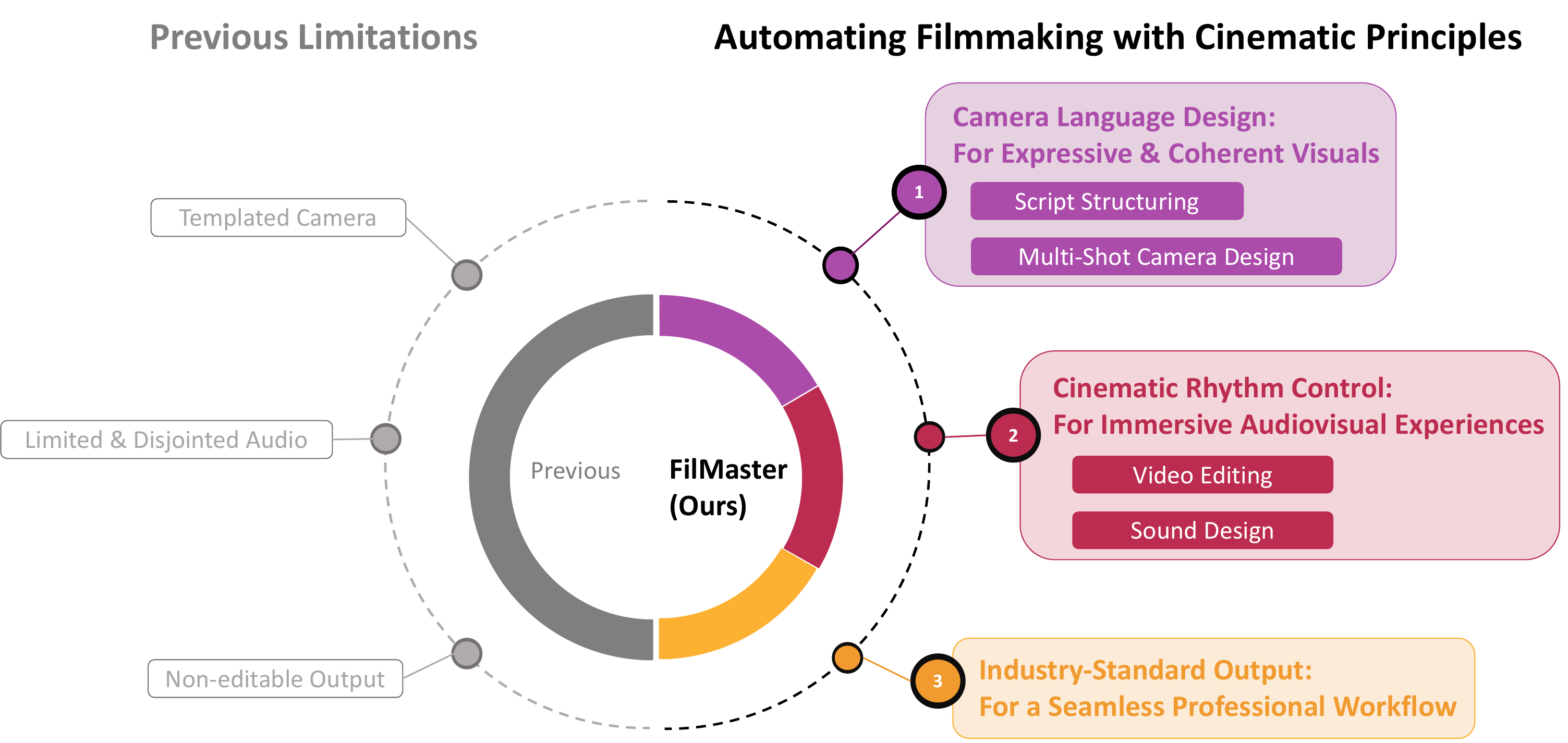}
    % \vspace{-10pt}
   \caption{Comparison of our \name{} with AI-driven current workflow and film generation system. 
   % \xuefei{"Utilize" real cine for expressive \& coherent visuals... "enagaging..." -> high-level single term, unify. "professoianl post production" -> "Audience-centric post-production".  }
   }

   \label{fig:intro}
   % \vspace{-15pt}
\end{figure}

To address these challenges, we propose \name{}, an automated film generation system designed to bridge the gap from script to screen with industry-standard output (\Cref{fig:intro}). Our approach is built upon two key cinematic principles: 
% (1) \textbf{Learning Cinematography from Real Film Data} to utilize established visual language. and (2) \textbf{Emulating Professional, Audience-centric Post-Production Workflows} to control narrative structure and pacing, and construct immersive audiovisual experiences. 
(1) \textbf{Learning Cinematography from Extensive Real-World Film Data}. 
% Camera language is crucial in filmmaking for emphasizing narrative objectives, building a compelling atmosphere for narrative coherence, and expressiveness. Filmmakers traditionally hone this skill by studying extensive film references~\cite{nelmes2012introductionFilm}. 
Filmmakers traditionally hone camera language skills, vital for narrative coherence and expressiveness, by studying extensive film references. Drawing inspiration from this practice, \name{} learns from a vast corpus of real film clips to master and apply professional camera language. 
(2) \textbf{Emulating Professional, Audience-Centric Post-Production Workflows}. 
This principle directly addresses the challenge of cinematic rhythm.
Post-production is crucial in filmmaking for an engaging experience, where proper pacing, and the effective integration of audiovisual elements are paramount~\cite{case2013filmPostProduction}. Our \name{} emulates the post-production workflow to control narrative pacing, and construct immersive audiovisual experiences. Inspired by the recent progress of MLLMs, which show great promise in understanding complex text scripts, videos, and audio, we introduce MLLMs to implement such cinematic principles in \name{}.

% \TODO{two stages clear writing, and highlight two modules in separation}
% Leveraging the advancing capabilities of (M)LLMs beyond scriptwriting into broader creative tasks in film generation, 
% % \name{} is architected around these powerful models.
% \name{} integrates cinematic principles through a two-stage process:
\name{} integrates cinematic principles for film generation through a two-stage process from generation to post-production, leveraging advanced (M)LLMs for tasks beyond scriptwriting:
% the \generation{} that transforms the user input (text prompt and reference images) into text-coherent and visually expressive video sequences, and the \coordination{} that orchestrates the generated videos and audio clips into cinematic outputs with engaging multimodal rhythms.
(1) \generation{}. %This initial stage transforms the user's input into visually rich video sequences. 
This stage takes the input text and the reference character/location images as input, leverages (M)LLMs for video content planning and script structuring, and then generates video clips with video generation models.
% Motivated by the importance of diverse and professional camera language in film production, 
Guided by our first principle, we propose \textbf{\camera{}} module to address the shortcomings of existing methods that often produce templated or overly generic camera language by learning professional cinematography from real films.
% While previous AI-based film generation approaches produce videos with templated and overly generic camera language, our framework follows the principle of learning professional cinematography techniques from reference films to foster coherent and expressive camera language design.
% By leveraging Retrieval-Augmented Generation (RAG) over a vast dataset of 440,000 real film clips, \name{} retrieves video clips based on indexed context in text embeddings. 
\name{} utilizes RAG with a vast dataset of 440,000 real film clips to retrieve relevant cinematic descriptions, which are indexed by film clips' annotated textual descriptions, based on the query of scene's textual context. To synergize the camera language between multiple shots within the same scene, we take multiple shots into consideration with the spatio-temporal contexts and narrative objectives to retrieve reference film clips.
%The meticulous spatio-temporal design of this input scene context promotes a highly coherent and expressive camera language within that scene, leading to synergized multi-shot outputs. 
These retrieved textual descriptions then serve as references for an LLM to learn professional cinematography to re-plan shots, including shot types, camera movements, angles, and atmospheric characteristics, aligned with the scene's narrative objectives. By grounding in professional films, the system is empowered to generate expressive, contextually relevant shots with coherent camera language, ensuring both cinematic-standard coherence and narrative expressiveness.
(2) \coordination{}. %The subsequent stage focuses on orchestrating the generated video and audio clips into a polished final output.
This stage transforms the raw footage from the previous stage into rich, multi-layered audiovisual outputs, orchestrating both visual and auditory elements to achieve cinematic rhythms.
%Our core innovation in this stage is\textbf{\rhythm{}}: 
Guided by our second principle, we propose \textbf{\rhythm{}} module to address the common issue of flat, unengaging AI-generated content by emulating professional, audience-centric post-production workflows. 
% \xuefei{somehow strange english} 
% Starting with the assembly of a \textit{Rough Cut} to establish the basic narrative structure, an MLLM-based audience-centric review mechanism generates feedback, which guides a sophisticated \textit{Fine Cut} editing process. This MLLM is provided with demographic profile that specifies certain audience type (\textit{e.g.} short-drama audience) by searching the Internet. Then the editing process enables the effective integration of audiovisual elements: Video editing involves detailed structural and durational adjustments to video clips. Sound design ensures that a rich, multi-layered soundscape is crafted, integrating diverse audio elements (ambiance, musical score, voice-overs(VO), foley, and SFX) with multi-scale audiovisual synchronization. 
% The audience-centric review, \textit{Rough Cut}, and \textit{Fine Cut} are all implemented  by MLLMs prompted to act as different roles (\ie, MLLMs as audience, professional film editors, and sound designers).
The process begins with assembling a \textit{Rough Cut} to establish the basic narrative structure. Next, an MLLM, acting as a simulated audience, reviews this \textit{Rough Cut} and generates feedback. This feedback then guides a sophisticated \textit{Fine Cut} editing process.
To tailor this review, the MLLM can be provided with a demographic profile,
% (\textit{e.g.}, ``short-drama audience'')
potentially sourced from internet searches, to simulate a specific target audience.
This feedback then guides a sophisticated \textit{Fine Cut} editing process, which focuses on the effective integration of audiovisual elements: 
Video editing involves detailed structural and durational adjustments to video clips. 
Sound design ensures that a rich, multi-layered soundscape is crafted, integrating diverse audio elements (background ambiance, musical scoring, voice-overs (VO), foley, and sound effects (SFX)) with multi-scale audiovisual synchronization. 
Notably, MLLMs drive this entire process, from audience-centric review to \textit{Rough} and \textit{Fine Cut} editing, by being prompted to adopt distinct professional roles (\textit{e.g.}, audience, film editor, sound designer). 
Through this audience-centric process, \name{} adjusts narrative pacing and enables effective integration of audiovisual elements, resulting in outputs with immersive and engaging rhythms. %, and ultimately ensures emotional resonance and audience engagement. %\xuefei{unify wording}

Furthermore, to address the practical limitations of prior systems, which often produce non-editable video files, limiting practical industry integration, \name{} generates editable and structured output videos including audio clips, with multi-track timelines using the industry-standard OpenTimelineIO (OTIO) format~\cite{openTimelineIO}. This allows seamless export to professional editing software like \textit{DaVinci Resolve}, directly bridging AI generation with professional film production workflows.

To address the lack of suitable benchmarks for AI film generation, we introduce \eval{}. Existing benchmarks~\cite{zheng2025vbench2, zhuang2025vistorybench} often focus solely on visual results, lacking capabilities for a holistic film evaluation. In contrast, \eval{} is a comprehensive benchmark that evaluates quality across key cinematic dimensions, covering narrative, audiovisual techniques, aesthetics, rhythm, engagement, and overall quality. Our experiments demonstrate \name{}'s superior performance, particularly in sophisticated camera language design and cinematic rhythm control.

To summarize, our contributions are as follows: 
% \begin{itemize}
% \item We propose \name{}, the first comprehensive AI-based film generation system that is explicitly designed around cinematic principles to guide camera language and cinematic rhythm.
% % of learning cinematography from real films and emulating professional, audience-centric post-production workflows.
% The system bridges the gap from script to screen by incorporating \generation{} and \coordination{}, and produces editable and structured output, ensuring integration with real-world film production workflows.
% \item We introduce a novel \camera{} module that generates coherent, expressive visuals by learning cinematography from a vast corpus of real films, leading to synergized multi-shot outputs.
% \item We present an innovative \rhythm{} module. This module emulates professional post-production, leveraging MLLMs to control narrative structure and pacing, and integrates rich, multi-layered audiovisual elements from audience-centric review for immersive and engaging output.
% \item We propose a comprehensive benchmark (\eval{}) and evaluation validating \name{}'s superior capabilities in generating high-quality and engaging cinematic content.
% \end{itemize}
\begin{itemize}
    \item \textbf{A Novel System Integrating Cinematic Principles.} We propose \name{}, the first comprehensive AI-based film generation system explicitly designed around cinematic principles to guide camera language and cinematic rhythm. It bridges the gap from script to screen and produces editable, structured output compatible with professional production workflows.

    \item \textbf{Learning Cinematography from Real Films.} We introduce a novel \camera{} module that generates coherent and expressive visuals by learning cinematographic patterns from a vast corpus of real films, resulting in synergized multi-shot outputs.

    \item \textbf{AI-Driven Post-Production for Cinematic Rhythm.} We present an innovative \rhythm{} module that emulates professional post-production. It leverages MLLMs to control the narrative pacing, and integrates audiovisual elements based on audience-centric analysis for immersive experiences.

    \item \textbf{A Comprehensive Film Evaluation Benchmark.} We establish a new benchmark, \eval{}, for the holistic evaluation of AI-generated films and provide experiments that validate \name{}'s superior performance in creating high-quality, engaging cinematic content.
\end{itemize}

\begin{table}[t]
\centering
\caption{
Comparison of film generation capabilities across different AI methods: Anim-Director~\cite{li2024animdirector}, MovieAgent~\cite{wu2025movieagent}, and LTX-Studio (commercial product)~\cite{LTXStudioAppMisc}.
}
\label{tab:intro_compare}
\vspace{-0.2cm}
\resizebox{\linewidth}{!}{
\begin{tabular}{l cccccccc} % l for first column, c for the rest
\toprule
\multirow{2}{*}{Method} & \makecell{Script \\ Design} & \makecell{Camera \\ Language \\ Design} & \makecell{Audio \\ Types} & \makecell{A/V \\ Sync} & \makecell{Video Adj. \\ (Structure)} & \makecell{Video Adj. \\ (Duration)} & \makecell{Audience \\ Review} & \makecell{Editable \\ Output} \\
% & & & & & & & & \\ % This empty row is for the multirow spanning
% \cmidrule(lr){1-1} \cmidrule(lr){2-9} % Simpler cmidrules for this header style
% Or even simpler, just one \cmidrule(lr){2-9} if Method is clearly separate
% Or, remove cmidrules entirely if you prefer the absolute minimal look from your example's data rows

\midrule % Main separator after header

% Academic Work
% \makecell[l]{Academic Work \\ (e.g., MovieAgent, \\ AnimDirector)} & 
% \cmark & 
% No/Template & 
% 0--1 & 
% \xmark & 
% \xmark & 
% \xmark & 
% \xmark & 
% \xmark \\
\makecell[l]{Anim-Director} & 
\cmark & 
\xmark & 
0 & 
\xmark & 
\xmark & 
\xmark & 
\xmark & 
\xmark \\

\makecell[l]{MovieAgent} & 
\cmark & 
Templated & 
1 & 
\xmark & 
\xmark & 
\xmark & 
\xmark & 
\xmark \\
% \midrule % Optional: lighter rule or no rule between these rows for minimalism

% Commercial Product
\makecell[l]{LTX-Studio} & 
\cmark & 
Templated & 
1 & 
Limited & 
\xmark & 
\xmark & 
\xmark & 
\cmark \\
% \midrule % Optional

% Ours
\textbf{\name{} (Ours)} & 
\cmark  & 
\cmark (Film-based) & 
5 & 
\cmark & 
\cmark & 
\cmark & 
\cmark & 
\cmark \\
\bottomrule
\end{tabular}
}
\end{table}

% P2 concrete problems; we propose system to adress

% P3 highlight two modules

% P4 eval

% P5 contributions

\section{Related Work}

\subsection{Video Generation}
% ref movieagent and T2V-compbench

% introduce diffusion model and language based models --> our work differs from that in generating whole film

Existing video generation models can be roughly categorized into diffusion model-based~\cite{ho2022imagen, singer2022make, zhou2022magicvideo, khachatryan2023text2video, luo2023videofusion, blattmann2023align, he2022latent, wang2023modelscope, yang2024cogvideox}, and 
language model-based~\cite{villegas2022phenaki, chang2023muse, kondratyuk2023videopoet, yu2023magvit, yu2023language, chang2022maskgit}. Video diffusion models excel by progressively refining noisy inputs into clean video samples, with recent advancements like Sora~\cite{videoworldsimulators2024}, HunyuanVideo~\cite{kong2024hunyuanvideo}, and Wan-Video~\cite{wan2025} demonstrating remarkably high-quality visual synthesis via sophisticated latent diffusion techniques. 
Video language models, such as VideoPoet~\cite{kondratyuk2023videopoet}, are typically derived from the family of transformer-based language models that can flexibly incorporate multiple tasks in pretraining, and show powerful zero-shot capabilities.

% \subsection{Story visualization}
% \TODO{to add}
% % % 早期：Story visualization
% % % our work differs in 呈现方式

\subsection{LLMs for Film Production}
% anim-director, movieagent, FilmAgent
% our work differs from them in end-to-end generation, camera languange and cinematic rhythm control by referencing professional film clips and workflows.
Recent works for preliminary exploration in film production have begun to leverage the emerging reasoning and planning capabilities of LLMs~\cite{wei2022emergent, Lin2023VideoDirectorGPT}. Anim-Director~\cite{li2024animdirector} automates animation generation by employing large multimodal models (LMMs) to expand user inputs into coherent storylines, generate scene images and videos by iteratively optimization.
The development of multi-agent systems~\cite{park2023generativeagentsinteractivesimulacra, sun2024corexpushingboundariescomplex, wu2023autogenenablingnextgenllm,hong2024metagptmetaprogrammingmultiagent, yuan2024moraenablinggeneralistvideo, huang2024genmac} has further spurred innovation. For example, FilmAgent~\cite{xu2025filmagent} proposes multi-agent collaboration through iterative feedback and revisions in constructed 3D virtual spaces.
MovieAgent~\cite{wu2025movieagent} multi-agent Chain of Thought planning for movie generation, which automatically structures scenes, camera settings, and cinematography. 
Distinct from these prior efforts, \name{} significantly expands the role of LLMs to orchestrate a more comprehensive end-to-end film production pipeline, from pre-production through post-production. Crucially, our system uniquely integrates the learning of camera language from real film footage and the emulation of professional post-production workflows for cinematic rhythm control, aspects not holistically addressed by previous LLM-based film generation systems (\Cref{tab:intro_compare}).

\section{Method}
We introduce the overview of our system in~\Cref{subsec: overview}, and detail the two core innovation modules, \camera{} module in~\Cref{subsec: camera language design} and \rhythm{} module in~\Cref{subsec: cinematic rhythm}, respectively. 

\begin{figure*}[ht]   % htbp
  \centering
    % \fbox{\rule{0pt}{1.5in} \rule{1\linewidth}{0pt}}
   % \includegraphics[width=\linewidth]{figures/pipeline_v2.pdf}
   \includegraphics[width=\linewidth]{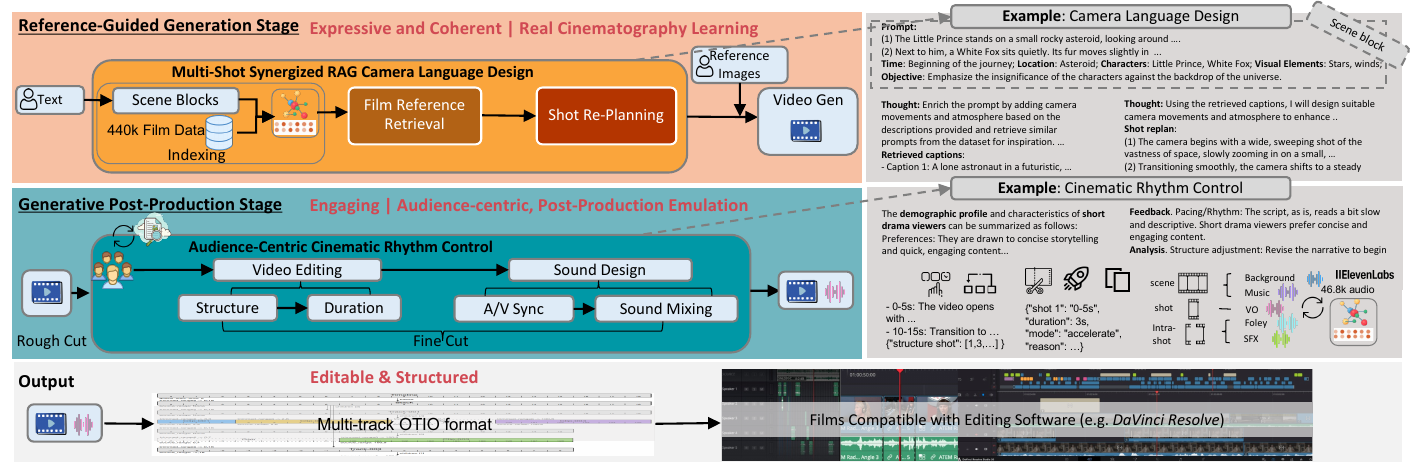}
   \vspace{-0.2cm}
   \caption{Overview of the \name{} framework for automated film generation. It processes user input to produce editable, structured outputs, guided by cinematic principles. Key innovations include the (1) \colorbox{mycolor_orange}{\camera{}} module within the \generation{} stage, leveraging real film data for coherent and expressive visuals, and the (2) \colorbox{mycolor_deepblue}{\rhythm{}} module within the \coordination{} stage, emulating professional post-production for enhanced audience engagement. Examples are shown in \colorbox{mycolor_gray}{gray} area.
   }

   \label{fig:pipeline}
   \vspace{-0.5cm}
\end{figure*}
% \subsection{Overview}
% motivation
% method
% (effect)
\subsection{Overview of \name{}}
\label{subsec: overview}

\name{} is an automated film generation system designed to produce complete films from an input text, guided by reference images for characters and locations, and generating outputs fully editable and structured, with multi-track timelines using the industry-standard OTIO format. As illustrated in~\Cref{fig:pipeline}, the overall process can be conceptualized into a two-stage process:
\begin{itemize}
    \item \generation{}: This stage takes the input text and the reference character/location images as input, leverages (M)LLMs for video content planning and script structuring, and then generates video clips with video generation models. This involves first, from coarse-to-fine, progressively refining the initial input text into detailed scene descriptions with spatio-temporal contexts. Subsequently, the camera language for each shot within the same scene is synergistically planned. Finally, video clips are generated based on this designed visual language and the reference images. Our \camera{} module (detailed in~\Cref{subsec: camera language design}) plays a crucial role in crafting coherent and expressive visual language.
    \item \coordination{}: Based on the generated videos from the \generation{}, this stage transforms the raw footage into polished final outputs, orchestrating both visual and auditory elements to achieve cinematic rhythms. This stage includes assembling a \textit{Rough Cut}, refining this cut based on simulated audience-centric feedback to achieve a \textit{Fine Cut}, with structural and durational adjustments in video editing and a multi-layered soundscape in sound design. Our \rhythm{} module (detailed in~\Cref{subsec: cinematic rhythm}) is central to controlling narrative structure and pacing, ensuring effective integration of audiovisual elements, and thereby improving emotional resonance and audience engagement.
    % \item Output Stage: In the final stage, the completed audio-visual timeline is packaged into the OpenTimelineIO (OTIO) format. This ensures compatibility with professional editing software, facilitating seamless integration into existing film production workflows.
\end{itemize}
The completed videos and audio are packaged into the industry-standard OTIO format with multi-track timeline. This ensures compatibility with professional editing software, facilitating seamless integration into real-world film production workflows.

% To address the limitations of templated camera language design, and the lack of cinematic rhythm mastery, \name{} incorporates: (1) Scene-Level, Multi-Shot Camera Language Design, and (2) Audience-Centric, Multimodal Cinematic Rhythm Control. The first component enables the creation of scene-coherent camera language by referencing real film footage within specific script contexts. The second facilitates engaging narrative and emotional pacing by emulating professional post-production workflows to orchestrate video clips and the overall audio landscape from the perspective of audience.

% \subsection{Scene-Level, Multi-Shot Camera Language Design}
% \subsection{\generation{}}
% \subsection{RAG-based Multi-shot Camera Language Design}
\subsection{\camera{}}
\label{subsec: camera language design}
% spatio-temporal contexts 
% rag
% \TODO{1. inspired by director xxx. 2. highlight benefit of ``Scene-Level, Multi-Shot Camera Language Design'' 3. explain overview and logic of RAG for cinematography.  }
Inspired by how professional filmmakers traditionally hone their camera language skills through extensive study of film references, \name{} incorporates a \textbf{\camera{}} module, referencing over a vast dataset of real film clips for camera language. This module overcomes the limitations of templated and overly generic camera work through multi-shot synergized RAG, which involves: (1) spatio-temporal-aware indexing for embedding scene contexts, (2) film reference retrieval for retrieving similar cinematic examples, and (3) shot re-planning for generating coherent and expressive cinematography.

% To overcome the limitations of templated and overly generic camera work, \name{} implements a principled approach to camera language design. This module, guided by the narrative contexts of the script, learns cinematographic patterns from real-world films to generate expressive and coherent multi-shot sequences at the scene level. The process involves:

% \textit{Script Structuring and Spatio-temporal Context Definition.}
\textbf{Spatio-Temporal-Aware Indexing.}
% spatio-temporal context
\name{} first processes the input text into scene blocks. 
A scene block defines a continuous segment of the narrative occurring within a single, coherent scene, thus maintaining continuity in both space and time for that specific scene. Each scene block is described by its spatio-temporal contexts, comprising elements of multi-shot prompts, the scene's location, time of day, present characters, and key visual elements, and its overarching narrative objective for that scene (example is shown in~\Cref{fig:pipeline} top-right part). Each shot within the same scene block shares the same reference images for characters and locations to maintain continuity.
This coarse-to-fine script structuring, employing a chain of LLMs, progressively refines the input text from a synopsis to a simple storyboard and a detailed storyboard and finally into scene blocks (detailed in~\Cref{app: method_scene_structuring}). 
% Crucially, each scene block contains spatio-temporal contexts (location, time, characters, key visual elements) and its narrative objective, which is important for continuity in the same scene. 
The meticulous design of the scene block with its spatio-temporal contexts and narrative objectives promotes a highly coherent and expressive camera language within that scene, leading to synergized multi-shot outputs. 
Secondly, the scene block is encoded into vector representations using an embedding model and stored in a vector database as text embeddings~\cite{gao2023rag, luo2024videoRAG}.
Each scene block, rich with spatio-temporal contexts and narrative objectives, serves as the precise query for the subsequent retrieval and generation process, ensuring that the learned camera language is deeply aligned with the specific narrative objectives of each shot within the same scene.
Our real film dataset contains over 440,000 professionally annotated film clips. These annotated textual descriptions detail key aspects of cinematic descriptions of camera language, including shot types, camera movements, angles, and atmospheric characteristics. The textual descriptions of the film dataset are also encoded into a vector representation using the same embedding model. 

% \textit{Retrieval-Augmented Generation (RAG) for Cinematography Learning.}
\textbf{Film Reference Retrieval.}
% This approach grounds our camera language design in professional cinematic techniques, moving beyond LLM-only imagination or rigid templates. 
The scene block, previously defined through Spatio-Temporal-Aware Indexing and incorporating its spatio-temporal contexts and narrative objective, serves as the query for retrieval.  Its vector representation (the query vector) is compared against the film dataset vectors, and similarity scores are computed. Then the process prioritizes and retrieves the top-$K$ film references that demonstrate the greatest similarity to this query vector. The text descriptions of the retrieved flim clips are subsequently used as references to guide the LLM-based shot re-planning in the next paragraph.

% \textit{Cinematographic Pattern Analysis and Shot Re-planning.}
\textbf{Shot Re-planning.}
Building upon the retrieved references, \name{} analyzes recurring cinematic patterns and extracts professional camera techniques applicable to the current narrative context. This analysis focuses on identifying effective visual storytelling methods that can enhance the visual impact and narrative objectives of each scene block. 
Concretely, the original scene block query and the retrieved film references are synthesized into a coherent prompt for the LLM. The LLM then re-plans the multi-shot prompts to ensure consistent camera language within a scene block. %Existing conversational history can also be integrated into this prompt, enabling effective multi-turn dialogue interactions. 
The re-plan process can be applied iteratively with the multi-turn dialogue of LLMs.
This multi-shot synergized design within the scene block, guided by the scene's narrative objectives and informed by real film references, ensures multi-shot continuity and coherence, which is a key distinction from prior work that often treats shots in isolation. The shot re-planning generation specifies appropriate shot types, camera movements, angles, and atmospheric descriptions for each shot, while keeping the original narrative content and objectives intact. This process ensures the final camera language is not only expressive but also coherent at the multi-shot scene level.

% \textit{Video Generation Based on Shot Re-planning.}
% The replanned shots, now represented as enhanced visual prompts with explicit cinematographic specifications, are subsequently fed into a video generation model. This model uses multi-image references—derived from the characters and location identified in the scene block—as conditioning inputs. 
% By combining these references with cinematographically-informed prompts, the system generates visually consistent output that maintains continuity across the entire film sequence.

% \subsection{Audience-Centric, Multimodal Cinematic Rhythm Control}
% \subsection{\coordination{}}
% \subsection{Audience-centric Cinematic Rhythm Control}
\subsection{\rhythm{}}
\label{subsec: cinematic rhythm}

% While our Camera Language Design module generates visually coherent scenes, directly concatenating these outputs would result in a film with flat visual rhythm and unengaging emotional impact due to the lack of suitable audio, falling far short of professional standards. 
% Relying solely on the camera language design would produce content that is unengaging for audiences, lacking narrative drive and emotional depth.
While our \camera{} module generates visually coherent scenes, relying solely on this visual output, without suitable narrative drive and effective integration of audiovisual elements, would result in a film with flat, unengaging generated content, failing to connect with audiences, and falling far short of professional standards.
Therefore, to address this, we propose a \textbf{\rhythm{}} module. 
Inspired by professional film post-production workflows~\cite{case2013filmPostProduction} that progressively refine rhythm, this module emulates this process by first assembling and reviewing a \textit{Rough Cut} from a simulated audience-centric perspective. It then proceeds to a \textit{Fine Cut}, where it orchestrates the visual narrative structure and pacing through video editing, and integrates a rich, multi-layered soundscape through sound design, making the film compelling in both its emotional resonance and audience engagement. MLLMs drive this entire process, from audience-centric review to \textit{Rough} and \textit{Fine Cut} editing, by being prompted to adopt professional roles in film post-production (\textit{e.g.}, audience, film editor, sound designer). 
% audience
% editing
% sync

\textbf{Audience-Centric Review.}
Traditional AI approaches often operate solely from a director's perspective, potentially limiting the film’s emotional resonance and engagement with the actual audience. To bridge this gap, \name{} introduces an audience-centric review mechanism. This approach integrates both the director's intended narrative expression with simulated audience expectations.
% audience demographic profile, feedback, refinement suggestion
Our \name{} first allows for the specification of a target audience archetype (\textit{e.g.}, ``short-drama audience''). An MLLM then utilizes Internet search tools to construct a demographic profile detailing this archetype's characteristics, preferences, and typical viewing expectations (such as a preference for concise storytelling or fast-paced engagement).
To facilitate the review, a \textit{Rough Cut} version is assembled for basic narrative structure. This currently involves combining the video sequences generated by the \camera{} module with LLM-generated audio textual descriptions (VO) for each scene block, serving as a temporary placeholder for the soundscape for easier understanding by the audience. An MLLM, informed by the constructed audience demographic profile, then critiques this version. Rather than modifying content directly, the MLLM identifies potential misalignments in areas such as structural flow, narrative pacing, scene transitions, and the consistency of the placeholder audio descriptions.
Following the audience-centric critique, a separate LLM served as the analysis module that bridges critique and implementation by systematically categorizing identified issues into three dimensions: structural organization, timing and duration, and audio coherence.
This analysis produces actionable recommendations that directly inform the subsequent processes, translating audience-centric feedback into concrete adjustments in a \textit{Fine Cut}, with video editing and sound design.

\textbf{Video Editing.}
% In Rough Cut, video clips are assembled in chronological order to establish the full narrative arc, ensuring that the basic storyline is complete and structurally coherent without yet emphasizing pacing, visual rhythm, or audience engagement.
In video editing process, deeper refinement is performed based on audience analysis. Based on the audience's feedback and text description of video, annotated with precise timecodes, our system employs LLMs prompted to act as professional film editors to systematically address narrative structure and pacing, addressing issues related to logical flow and information density, through two primary mechanisms: 
% Picture lock helps keep the audio in sync with the video
(1) Structural Reorganization: Rearranging or removing redundant shots to create a more logical and compelling progression of scenes, improving narrative structure and audience understanding. 
(2) Duration Adjustment: Modifying the temporal length of individual clips to control information exposure per scene for narrative pacing.
Three operations are considered: trimming (removing redundant visual content), acceleration (speeding up segments to fit pacing needs), and retention (preserving original timing when appropriate).
% (3) Audio Refinement: Preparing for fine-grained multi-track audio synchronization, aligned with adjusted shot durations. Detailed audio-visual synchronization techniques are elaborated in Step 6.
This editing process progressively aligns visual storytelling with narrative objectives and audience expectations, refining the narrative structure and pacing, and culminating in a picture lock, which is a finalized visual sequence ready for detailed sound design.

\textbf{Sound Design. }
% Audio Incorporation and A/V Synchronization; Sound Mixing
Prior AI systems often neglect or implement limited audio, as shown in~\Cref{tab:intro_compare}, failing to leverage the power of effective audiovisual elements to create atmosphere and emotionally engage audiences beyond visuals alone. Thus, we propose multi-scale A/V synchronization for effective integration of audiovisual elements, including background ambience, music scoring, VO, foley, and SFX, forming a multi-layer soundscape, followed by sound mixing.
Directly relying on MLLM to design multi-track audio (including background ambience, musical scoring, VO, foley, and SFX) often results in poor synchronization with video content. This is primarily due to the MLLM’s limited capability in managing the coordination of multiple audio layers requiring different temporal scales relative to the video. To address this, we propose a multi-scale audiovisual synchronization strategy, which systematically designs different audio types at appropriate temporal scales. This process is supported either by synthesized voice content for VO, or retrieval-augmented generation (RAG) from curated audio libraries for other audio types like background ambience, music scoring, foley and SFX (similar as in~\Cref{subsec: camera language design}).
% Specifically, synchronization is designed across three temporal scales, each use (M)LLM for analysis: 
% Scene-level synchronization: Background ambience and musical scoring are assigned at the scene level to establish
% atmosphere and emotional tone using the previous scene block. 
% Shot-level synchronization: VO, including narration and dialogue, is designed at the shot level to maintain close alignment with visual content and enhance narrative clarity.
% Intra-shot synchronization: Foley and SFX are assigned at a fine-grained temporal resolution. Agents analyze
% video content with second-level timecodes to detect specific actions, movements, and environmental elements that re-
% quire sonic representation, ensuring detailed audiovisual coordination.
Synchronization is managed across three key temporal scales: scene-level for broader elements like background ambiance and music scoring to establish overall atmosphere; shot-level for components like VO that align with specific shots; and intra-shot for fine-grained foley and SFX tied to precise actions or visual events within a shot. At the scene-level, an LLM directly utilizes the scene block to select appropriate music and background ambiance. For shot-level sound design, an LLM is informed by text descriptions of the video and audience feedback. In contrast, intra-shot level sound design employs an MLLM to achieve greater precision in aligning fine-grained audio cues with visual events.
% To ensure audio quality and contextual relevance, we employ the RAG approach for non-voice audio tracks. We construct a curated audio library comprising 46,826 sound assets, including 5,877 music tracks and 40,949 other audio assets (atmospheric sounds, foley, and effects) sourced from the Internet. Each asset is tagged with semantic descriptors, emotional qualities, and acoustic properties to facilitate context-aware retrieval. For voice-over content, we utilize ElevenLabs’ text-to-speech services, allowing for flexible and high-fidelity voice generation tailored to specific narrative demands.
% To ensure audio quality and relevance, non-voice audio tracks (ambiance, music, foley, SFX) are sourced using a RAG approach from a large curated library (>46k assets tagged with semantic and acoustic descriptors). Voice-over and dialogue are generated using high-fidelity text-to-speech services (e.g., ElevenLabs).
% Composing multiple audio tracks directly often leads to inconsistencies in loudness, frequency balance, and dynamic range. To mitigate these issues, we apply sound mixing techniques that ensure cross-track and track-video harmonization. This includes: 
% Loudness units relative to full scale (LUFS) normalization to maintain consistent perceived loudness across tracks, frequency adjustment to avoid spectral clashes between different types of audio, particularly ensuring voice intelligibility by attenuating competing frequencies in background tracks, and dynamic equalization to enhance clarity and cohesion.
Finally, to address inconsistencies in loudness, frequency balance, and dynamic range that arise when composing multiple audio tracks, \name{} applies automated sound mixing techniques (Detailed in~\Cref{app: sound_design}). These processes, including LUFS normalization and frequency adjustments, ensure cross-track harmonization, voice intelligibility, and overall sonic cohesion, resulting in a polished soundscape.

\section{Experiments}

\subsection{Experiment Setting}
\textbf{Implementation Details.}
We use \textit{GPT-4o}~\cite{gpt-4o} for script generation, RAG, video editing, and sound design (VO, background, music). For audience-centric review and sound design (foley and sound effect), we employ \textit{Gemini-2.0-Flash}~\cite{team2023gemini}. We use Kling Elements~\cite{kling} as the video generation model, 
% a proprietary transformer-based text-to-video diffusion model 
capable of incorporating multiple reference images as conditions. 
The generated video clips are at 1920×1080 resolution, comprising 153 frames per sequence.

\textbf{Evaluation Metrics.}
% (1) gemini (2) user study
% first define
As this work pioneers an end-to-end film generation task with comprehensive attention to camera language and cinematic rhythm, we establish \textit{\eval{}}, a holistic evaluation benchmark. \eval{} is based on six high-level dimensions essential for assessing film quality: \textit{Narrative and Script (NS)}, \textit{Audiovisuals and Techniques (AT)}, \textit{Aesthetics and Expression (AE)}, \textit{Rhythm and Flow (RF)}, \textit{Emotional and Engagement (EE)}, and \textit{Overall Experience (OE)}. 
% The first two dimensions address , while the latter four dimensions focus on cinematic rhythm. 
These dimensions are further decomposed into twelve specific criteria for detailed evaluation (criteria are detailed in Appendix~\Cref{app: evaluation_instructions}):
\begin{itemize}
    \item \textbf{NS}: Script Faithfulness (SF), Narrative Coherence (NC)
    \item \textbf{AT}: Visual Quality (VQ), Character Consistency (CC), Physical Law Compliance (PLC), Voice/Audio Quality (V/AQ)
    \item \textbf{AE}: Cinematic Techniques (CT), Audio-Visual Richness (AVR)
    \item \textbf{RF}: Narrative Pacing (NP), Video-Audio Coordination (VAC)
    \item \textbf{EE}: Compelling Degree (CD)
    \item \textbf{OE}: Overall Quality (OQ)
\end{itemize}

While our work highlights two key modules for camera language and cinematic rhythm, it's important to recognize that cinematic quality arises from the holistic integration of various elements. Our evaluation dimensions are therefore designed to capture not only the direct outputs of each module but also their synergistic impact on the final film: 
\begin{itemize}
\item The \camera{} module's impact is primarily evaluated through NS (SF, NC), ensuring visual storytelling aligns with the script, and through key visual aspects of AT (VQ, CC, PLC), reflecting the quality and coherence of the planned visual foundation. This module also lays the groundwork for effective AE (CT) by designing shots with inherent cinematic qualities and contributes to the visual component of AE (AVR).
\item The \rhythm{} module's effectiveness is measured by audio-related aspects of AT (V/AQ), the realized AE (CT, AVR) through sophisticated editing and sound design, the mastery of RF (NP, VAC), and the resulting EE (CD). This module coordinates the visual and auditory elements into a cohesive and impactful rhythmic experience, evaluated by the ultimate arbiter OE (OQ).
\end{itemize}

To evaluate our method, we employ both automatic evaluation metrics and user studies in \eval{}. Given the absence of existing automatic metrics tailored for this task, we propose \textit{Gemini-1.5-Flash} as the evaluation model designed to assess generated films across the defined dimensions. To ensure reliability, we validate the effectiveness of automatic evaluation by measuring its correlation with human judgments.

\textbf{Test Dataset.}
Our evaluation employs a diverse set of 20 test cases comprising two distinct prompt types: 10 cases from MoviePrompts~\cite{wu2025movieagent},
% (including 8 prompts derived from well-known movies and 2 from annotators),
which feature extensive and detailed descriptions, with an average of 100.4 words; and 10 shorter, more concise prompts with an average of 15.2 words, specifically designed by annotators to evaluate our method's flexibility in handling varied input complexities. 

\textbf{Comparing Models.} 
We compare our method against previous works on automatic film generation: 
animation generation method (Anim-Director~\cite{li2024animdirector}), 
movie generation method (MovieAgent~\cite{wu2025movieagent}),
and a commercial product (LTX-Studio~\cite{LTXStudioAppMisc}). 
Since LTX-Studio supports automatic SFX, we apply the same to ensure fair comparison. 
% Due to the high cost of the commercial product, we randomly selected 3 cases from the test dataset for evaluation. 
% For consistency, all methods are evaluated on both audio and non-audio scenarios\footnote{For methods not supporting audio generation, silent clips are used.}.
% \TODO{The first is without audio... add audio description? check movie gen can or not}

% \textbf{User Preference Study.}

\subsection{Quantitative Results}
\textbf{Automatic Evaluation.} 
The results are shown in~\Cref{tab:auto_eval}, with an average improvement of 58.06\% in our \name{}: 43.00\% in camera language and 77.53\% in cinematic rhythm, respectively.
Our analysis reveals that existing approaches like Anim-Director~\cite{li2024animdirector} and MovieAgent~\cite{wu2025movieagent} significantly underperform across multiple dimensions including NS, AE, RF, EE, and OE. These methods demonstrate particularly severe deficiencies in audio quality and video-audio coordination. In contrast, our proposed method achieves substantial improvements across all evaluation dimensions in \eval{}, with an average performance increases of 75\% and 69\% compared to Anim-Director and MovieAgent, respectively. When compared with the commercial product LTX-Studio, we observe that LTX-Studio struggles with script faithfulness, narrative coherence, narrative pacing, and audio quality, likely due to insufficient integration of camera language and audiovisual elements. Our approach outperforms LTX-Studio with an average improvement of 19.84\%, demonstrating the effectiveness of our film generation system.

\begin{table*}[t]
\centering
\caption{
Automatic evaluation of film production generation across 12 evaluation criteria on \eval{}, and 2 derived evaluation criteria for camera language (CL) and cinematic rhythm (CRh).
\textsuperscript{*} denotes commercial product.
\colorbox{mycolor_blue}{Blue} represents the derived metric for CL, and \colorbox{mycolor_green}{green} stands for CRh, applicable to the following~\Cref{tab:human_eval_concise}.
}
\label{tab:auto_eval}
\vspace{-0.2cm}
\resizebox{\linewidth}{!}{
% Format: l for Method, then 12 'c' for data, then ':' for dashed line, then 2 'c' for derived, then 'c' for Avg
% Total columns for data = 2 (NS) + 4 (AT) + 2 (AE) + 2 (RF) + 1 (EE) + 1 (OE) = 12
% Then separator, then 2 (Derived), then 1 (Avg)
% The column specifier string:
% l c c  c c c c  c c  c c  c  c : c c  c

% \begin{tabular}{l c c  c c c c  c c  c c  c  c c c  c}
\begin{tabular}{@{}l cc cccc cc cc c c ccc@{}}
% Using @{} to remove extra space at the beginning and end of the tabular, standard with booktabs
% *{n}{c} is a shorthand for n columns of type c
\toprule
\multirow{2}{*}{Method} & \multicolumn{2}{c}{NS $\uparrow$} & \multicolumn{4}{c}{AT $\uparrow$} & \multicolumn{2}{c}{AE $\uparrow$} & \multicolumn{2}{c}{RF $\uparrow$} & EE $\uparrow$ & OE $\uparrow$ & \multicolumn{2}{c}{Derived $\uparrow$ } & \multirow{2}{*}{Avg $\uparrow$} \\
\cmidrule(lr){2-3} \cmidrule(lr){4-7} \cmidrule(lr){8-9} \cmidrule(lr){10-11} \cmidrule(lr){12-12} \cmidrule(lr){13-13}
% \cmidrule(lr){14-15} % cmidrule for Derived cols (15th and 16th actual data cols after Method)
 & SF & NC & VQ & CC & PLC & V/AQ & CT & AVR & NP & VAC & CD & OQ & \cellcolor{mycolor_blue}{CL}& \cellcolor{mycolor_green}{CRh} & \\
\midrule
Anim-Director & 1.60 & 2.20 & 4.20 & 3.45 & 3.55 & 1.00 & 3.05 & 2.50 & 2.10 & 1.00 & 2.45 & 2.30 & \cellcolor{mycolor_blue}{2.96} & \cellcolor{mycolor_green}{1.94} & 2.45 \\
MovieAgent & 1.50 & 1.60 & 4.10 & 3.40 & 3.40 & 1.00 & 2.70 & 2.20 & 1.60 & 1.00 & 2.20 & 2.20 & \cellcolor{mycolor_blue}{2.74} & \cellcolor{mycolor_green}{1.74} & 2.24 \\
LTX-Studio\textsuperscript{*} & 2.50 & 3.00 & 4.95 & 4.10 & 3.90 & 3.10 & \textbf{4.10} & 3.85 & 3.15 & 4.10 & 3.65 & 3.75 & \cellcolor{mycolor_blue}{3.74} & \cellcolor{mycolor_green}{3.62} & 3.68 \\
\textbf{Ours} & \textbf{3.90} & \textbf{4.60} & \textbf{5.00} & \textbf{5.00} & \textbf{4.40} & \textbf{3.80} & \textbf{4.10} & \textbf{4.10} & \textbf{4.40} & \textbf{5.00} & \textbf{4.20} & \textbf{4.40} & \textbf{\cellcolor{mycolor_blue}{4.50}} & \textbf{\cellcolor{mycolor_green}{4.32}} & \textbf{4.41} \\
\bottomrule
\end{tabular}
}
\vspace{-5pt}
\begin{flushleft}
\footnotesize
Derived metrics (CL, CRh) are computed from preceding base metrics:  CL is calculated as the average of (SF, NC, VQ, CC, PLC) + 0.5 * average of (CT, AVR), CRh is calculated as the average of (V/AQ, NP, VAC, CD, OQ) + 0.5 * average of (CT, AVR).
\end{flushleft}
\vspace{-8pt}
\end{table*}

\textbf{User Study.} 
In addition to quantitative analysis, we conduct a user study to evaluate the quality of generated films. 
Five participants were asked to rate each video independently based on the criteria defined in \eval{}. 
We randomly selected five cases from our dataset, compared our \name{} with the other three methods. 
In total, we collected 1,200 ratings, with 100 votes per evaluation criteria.
We show the six dimensions in~\Cref{tab:human_eval_concise}, and detailed results in~\Cref{tab:human_eval_detail}.
The results show that our \name{} demonstrates superior performance in film generation compared to the prior methods, with an average increase of 68.44\% (70.65\% in camera language, and 65.61\% in cinematic rhythm).
%. %avg of (top improvement + bottom improvement (two methods avg))
\begin{table}[htbp]
\centering

\caption{User study on film generation methods. }
\label{tab:human_eval_concise} % Changed label slightly to avoid conflict if you keep both
\vspace{-0.2cm}
\resizebox{\linewidth}{!}{
\begin{tabular}{l ccccccccc} % l for left-aligned labels, r for right-aligned numbers
\toprule
Method & NS $\uparrow$ & AT $\uparrow$ & AE $\uparrow$ & RF $\uparrow$ & EE $\uparrow$ & OE $\uparrow$ &  \cellcolor{mycolor_blue}{CL $\uparrow$}& \cellcolor{mycolor_green}{CRh $\uparrow$} &Avg $\uparrow$ \\
\midrule

% \textbf{Ours\textsuperscript{*}} & \textbf{3.90} & \textbf{4.00} & \textbf{4.10} & \textbf{3.90} & \textbf{4.40} & \textbf{4.00} & \textbf{4.02} \\
Anim-Director & 1.94 & 2.16 & 1.94 & 2.12 & 2.12 & 2.36 &\cellcolor{mycolor_blue}{2.15} &\cellcolor{mycolor_green}{2.04} & 2.09 \\
MovieAgent & 1.57 & 1.63 & 1.70 & 1.70 & 2.20 & 2.27 & \cellcolor{mycolor_blue}{1.66}& \cellcolor{mycolor_green}{1.83} & 1.74 \\
LTX-Studio\textsuperscript{*}  & 2.28 & 3.04 & 3.22 & 2.90 & 3.07 & 3.00 & \cellcolor{mycolor_blue}{2.80} & \cellcolor{mycolor_green}{3.05} &2.92 \\ 
\textbf{Ours} & \textbf{3.70} & \textbf{3.80} & \textbf{3.80} & \textbf{3.73} & \textbf{3.93} & \textbf{3.87} &\cellcolor{mycolor_blue}{\textbf{3.76}}& \cellcolor{mycolor_green}{\textbf{3.82}} & \textbf{3.79} \\
\bottomrule
\end{tabular}
}
\end{table}

\textbf{Human Correlation.}
To validate our proposed automatic evaluation metrics, we measured their correlation with human judgments using Pearson's $r$, Kendall’s $\tau$ and Spearmanr’s $\rho$ in~\Cref{tab:human_corr_concise}, similar to the methods in~\cite{park2021benchmark, huang2023t2i-compbench}. The automatic metrics exhibited an average correlation of 0.6230 with the user study results, indicating a strong alignment with human assessments.
\begin{table}[htbp] 
\centering

\caption{Human correlation coefficients of the average across 12 evaluation criteria in terms of Pearson Correlation $r$, Spearman's $\rho$, and Kendall's $\tau$ Coefficient ($p$-value < 0.01). }
\label{tab:human_corr_concise}
\vspace{-0.2cm}
\resizebox{\linewidth}{!}{
\begin{tabular}{lccc}
\toprule
Corrrelation          & Pearson $r$ & Spearman $\rho$ & Kendall $\tau$ \\
\midrule
Avg & 0.6464  & 0.6493          & 0.5732         \\
\bottomrule
\end{tabular}
}
\vspace{-1.0em}
\end{table}

\subsection{Qualitative Results}
% 展示结果（小王子、Nezha、Nemo turtle、Malena）+对比其他方法
% 对比的结果可以放到正文的最后两页
% 展示结果：script，image reference，video，audio

\begin{figure*}[ht]   % htbp
  \centering
    % \fbox{\rule{0pt}{1.5in} \rule{1\linewidth}{0pt}}
   % \includegraphics[width=\linewidth]{figures/multi_agent_v5.pdf}
   \includegraphics[width=\linewidth]{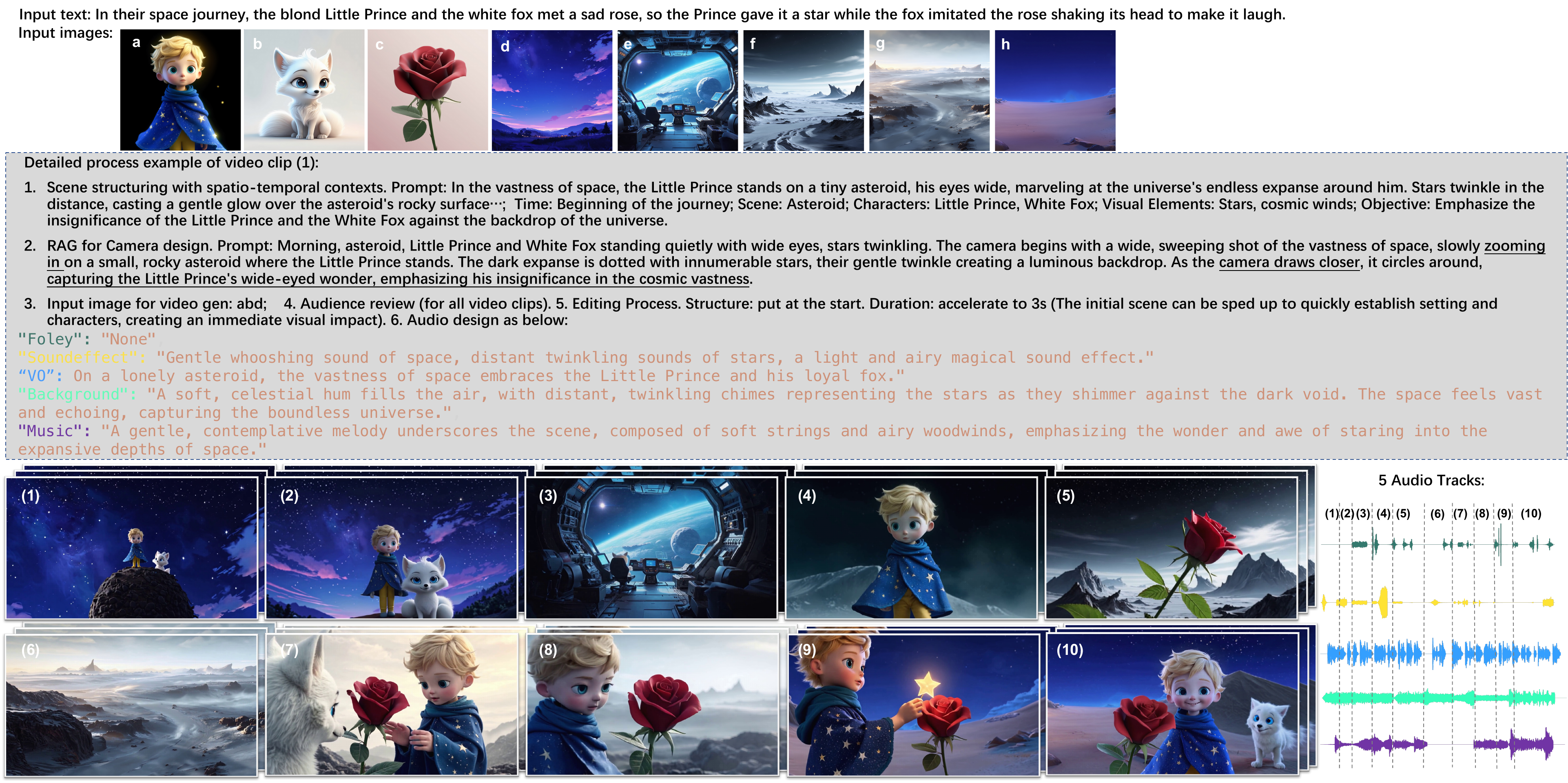}
    % \vspace{-10pt}
   \caption{The input and corresponding video frame extractions from our \name{}, illustrated alongside associated multi-track audio. The detailed process of the video clip (1) is illustrated in \colorbox{mycolor_gray}{gray} area.}
   \label{fig:result1}
   % \vspace{-15pt}
\end{figure*}

\textbf{Example.}
% script-camera-audio design, video is after orchestratioin
As shown in~\Cref{fig:result1}, our method generates descriptions with camera language and designs a multi-track audio based on the text prompt derived from the input text, forming a cohesive audiovisual narrative after camera language design and cinematic rhythm control. 
Additional examples are provided in~\Cref{fig:combined_results}.

\begin{figure*}[ht]   % htbp
  \centering
    % \fbox{\rule{0pt}{1.5in} \rule{1\linewidth}{0pt}}
   % \includegraphics[width=\linewidth]{figures/multi_agent_v5.pdf}
   \includegraphics[width=\linewidth]{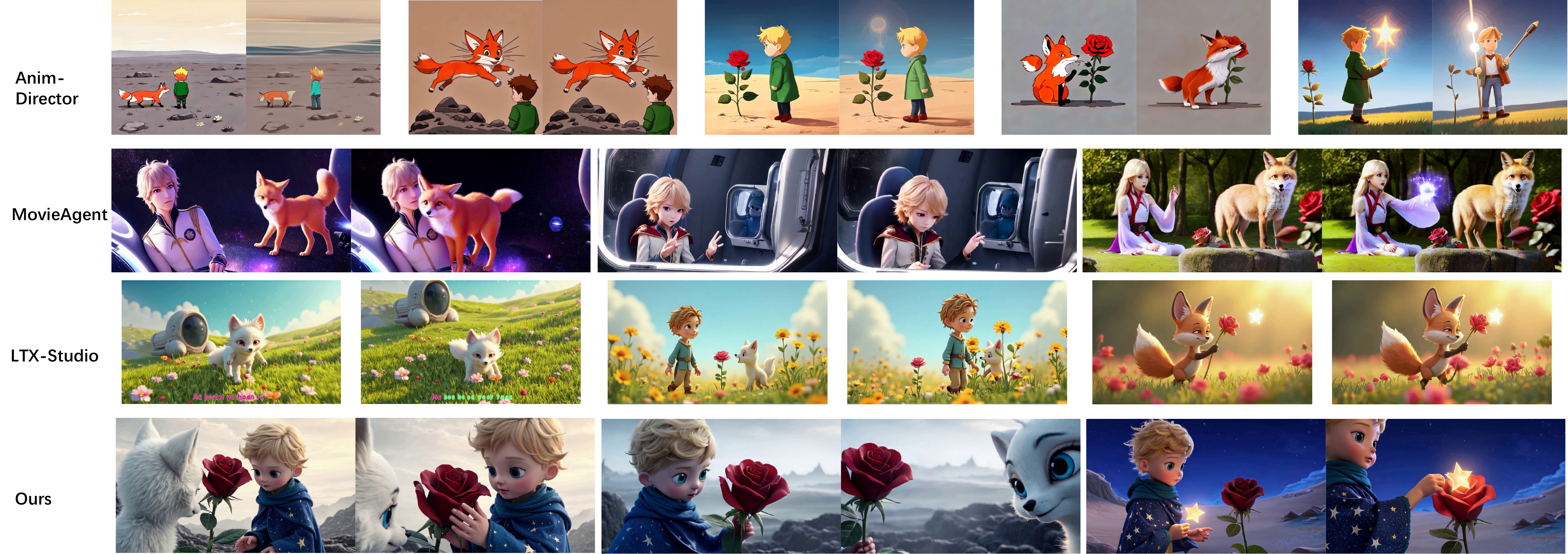}
    % \vspace{-10pt}
   \caption{
   Qualitative comparison of film production generation in the case of ``little prince''. Anim-Director, MovieAgent, and LTX-Studio adopt a text-to-image (T2I) followed by image-to-video (I2V) pipeline, where the input is script text only, without any images.
   }
   \label{fig:compare1}
   % \vspace{-15pt}
\end{figure*}

\textbf{Comparison.}
% Anim-Director: PPT-style
% MovieAgent: not consistency
% above they both struggle with video quality. no audio (MovieAgent provide simple VO for video)
% LTX-Studio: 
% good: visual quality, but struggle with narration
% narrative pacing slow and repetitive; desynchronization; not consistency in ID
The comparison is illustrated in~\Cref{fig:compare1}. Among all compared methods, our approach generates results with character consistency, fluid motion, and coherent narrative structure. In contrast, existing methods exhibit various limitations in visual quality, audio design, and narrative coherence: Visually, Anim-Director produces static animations that lack natural motion transitions. MovieAgent faces challenges with character consistency throughout the video. LTX-Studio, while achieving good visual quality, struggles with maintaining character identity preservation across frames.
Regarding audio and narrative elements, significant limitations exist across all compared methods. Anim-Director lacks audio altogether, severely restricting its storytelling capabilities. MovieAgent only implements basic voice-over without diverse audio design. LTX-Studio relies on automatic audio design without fine-grained control, resulting in desynchronization between visual and audio components. Furthermore, LTX-Studio's narrative pacing is often slow and repetitive. 
% These combined limitations in visual coherence, audio design, and narrative structure significantly restrict the expressive capabilities of existing methods, often resulting in misaligned audio-visual storytelling and tedious extended sequences that fail to maintain viewer engagement.

\subsection{Ablation Study}
% \textbf{Scene Structuring.}
% \textbf{RAG for Video Generation.}
% \textbf{Audience Perspective.}
% \textbf{Editing Strategy.}
% \textbf{Audio-Visual Synchronization.}
% \textbf{Audio Composition.}

% scene structuring, rag for video, audio co-design leads to higher performance compared to baseline (4.0 v.s. 3.5)
% orchestration (4.6 v.s. 4.0) for content expression.
\begin{table}[t]
\vspace{-0.2cm}
\centering
\caption{
Ablation study on the effects of Camera Language Design Module and Cinematic Rhythm Control Module on overall performance (average across 12 criteria). 
}
\label{tab:ablation_study_concise}
\resizebox{\linewidth}{!}{
\begin{tabular}{lccc}
\toprule
Method & w/o Camera + Rhythm & w/o Rhythm & Ours \\
\midrule
Avg $\uparrow$ & 3.75 & 4.17 & \textbf{4.67} \\
\bottomrule
\end{tabular}
}
\vspace{-1.5em}
\begin{flushleft}
% \footnotesize
% \textbf{Note:} Higher is better. This ablation indicates that both co-design and orchestration contribute to performance improvements.
\end{flushleft}
\end{table}

We ablate our method without \camera{} module and \rhythm{} module in~\Cref{tab:ablation_study_concise} on one case. Quantitative results show that removing cinematic rhythm module significantly decreases the average score in \eval{}, highlighting its role in cinematic expression with similar generated content. We also examined the impact of \camera{} module, which helps to form a coherent generated content.

% \begin{figure*}[htbp] % Using htbp for more placement flexibility
%   \centering
  
%   % First image
%   \includegraphics[width=\linewidth]{figures/result2.pdf}
%   % \vspace{1ex} % Optional: add a little vertical space between images if needed
  
%   % Second image
%   \includegraphics[width=\linewidth]{figures/result3.pdf}
  
%   % \vspace{-10pt} % Optional: adjust space before caption if needed
%   \caption{
%   Qualitative results of our \name{}. Our \name{} can generate high-quality films with either simple or complex script input.
%   }

%   \label{fig:combined_results} % Use a new, unique label for the combined figure
%   % \vspace{-15pt} % Optional: adjust space after figure if needed
% \end{figure*}

\section{Conclusion}
We introduced \name{}, the first comprehensive AI-based
film generation system designed for professional-grade film generation. \name{} uniquely integrates cinematic principles, focusing on camera language design and cinematic rhythm control, while ensuring industry-compatible, editable outputs.
We propose a \camera{} module that learns cinematography directly from a vast corpus of 440,000 real film clips. By leveraging Retrieval-Augmented Generation (RAG), this module produces expressive, context-aware camera plans with high cinematic coherence.
our \rhythm{} module emulates professional post-production. This includes \textit{Rough Cut} assembly, a \textit{Fine Cut} process refined by simulated audience feedback, including video editing and sound design, all orchestrated to achieve compelling narrative flow and profound emotional impact.
Furthermore, we proposed \eval{}, a comprehensive benchmark for assessing AI-generated films across six key cinematic dimensions. 
Extensive experiments demonstrate \name{}'s state-of-the-art performance, with an average improvement of 68.44\% in user studies and 58.06\% in automatic evaluation compared to prior methods, showcasing significant advancements in generating films with expressive visual language and engaging rhythm. 

% Bibliography
% \clearpage
{
    \small
    \bibliographystyle{ieeenat_fullname}
    \bibliography{main}

\begin{thebibliography}{49}
\providecommand{\natexlab}[1]{#1}
\providecommand{\url}[1]{\texttt{#1}}
\expandafter\ifx\csname urlstyle\endcsname\relax
  \providecommand{\doi}[1]{doi: #1}\else
  \providecommand{\doi}{doi: \begingroup \urlstyle{rm}\Url}\fi

\bibitem[Arijon(1976)]{arijon1976grammar}
Daniel Arijon.
\newblock Grammar of the film language.
\newblock \emph{(No Title)}, 1976.

\bibitem[Blattmann et~al.(2023)Blattmann, Rombach, Ling, Dockhorn, Kim, Fidler, and Kreis]{blattmann2023align}
Andreas Blattmann, Robin Rombach, Huan Ling, Tim Dockhorn, Seung~Wook Kim, Sanja Fidler, and Karsten Kreis.
\newblock Align your latents: High-resolution video synthesis with latent diffusion models.
\newblock In \emph{Proceedings of the IEEE/CVF Conference on Computer Vision and Pattern Recognition}, pages 22563--22575, 2023.

\bibitem[Bordwell et~al.(2004)Bordwell, Thompson, and Smith]{bordwell2004filmart}
David Bordwell, Kristin Thompson, and Jeff Smith.
\newblock \emph{Film art: An introduction}.
\newblock McGraw-Hill New York, 2004.

\bibitem[Brooks et~al.(2024)Brooks, Peebles, Holmes, DePue, Guo, Jing, Schnurr, Taylor, Luhman, Luhman, Ng, Wang, and Ramesh]{videoworldsimulators2024}
Tim Brooks, Bill Peebles, Connor Holmes, Will DePue, Yufei Guo, Li Jing, David Schnurr, Joe Taylor, Troy Luhman, Eric Luhman, Clarence Ng, Ricky Wang, and Aditya Ramesh.
\newblock Video generation models as world simulators.
\newblock 2024.

\bibitem[Case(2013)]{case2013filmPostProduction}
Dominic Case.
\newblock \emph{Film technology in post production}.
\newblock Routledge, 2013.

\bibitem[Chang et~al.(2022)Chang, Zhang, Jiang, Liu, and Freeman]{chang2022maskgit}
Huiwen Chang, Han Zhang, Lu Jiang, Ce Liu, and William~T Freeman.
\newblock Maskgit: Masked generative image transformer.
\newblock In \emph{Proceedings of the IEEE/CVF Conference on Computer Vision and Pattern Recognition}, pages 11315--11325, 2022.

\bibitem[Chang et~al.(2023)Chang, Zhang, Barber, Maschinot, Lezama, Jiang, Yang, Murphy, Freeman, Rubinstein, et~al.]{chang2023muse}
Huiwen Chang, Han Zhang, Jarred Barber, AJ Maschinot, Jose Lezama, Lu Jiang, Ming-Hsuan Yang, Kevin Murphy, William~T Freeman, Michael Rubinstein, et~al.
\newblock Muse: Text-to-image generation via masked generative transformers.
\newblock \emph{arXiv preprint arXiv:2301.00704}, 2023.

\bibitem[Gao et~al.(2023)Gao, Xiong, Gao, Jia, Pan, Bi, Dai, Sun, Wang, and Wang]{gao2023rag}
Yunfan Gao, Yun Xiong, Xinyu Gao, Kangxiang Jia, Jinliu Pan, Yuxi Bi, Yixin Dai, Jiawei Sun, Haofen Wang, and Haofen Wang.
\newblock Retrieval-augmented generation for large language models: A survey.
\newblock \emph{arXiv preprint arXiv:2312.10997}, 2\penalty0 (1), 2023.

\bibitem[He et~al.(2022)He, Yang, Zhang, Shan, and Chen]{he2022latent}
Yingqing He, Tianyu Yang, Yong Zhang, Ying Shan, and Qifeng Chen.
\newblock Latent video diffusion models for high-fidelity long video generation.
\newblock \emph{arXiv preprint arXiv:2211.13221}, 2022.

\bibitem[Ho et~al.(2022)Ho, Chan, Saharia, Whang, Gao, Gritsenko, Kingma, Poole, Norouzi, Fleet, et~al.]{ho2022imagen}
Jonathan Ho, William Chan, Chitwan Saharia, Jay Whang, Ruiqi Gao, Alexey Gritsenko, Diederik~P Kingma, Ben Poole, Mohammad Norouzi, David~J Fleet, et~al.
\newblock Imagen video: High definition video generation with diffusion models.
\newblock \emph{arXiv preprint arXiv:2210.02303}, 2022.

\bibitem[Hong et~al.(2024)Hong, Zhuge, Chen, Zheng, Cheng, Zhang, Wang, Wang, Yau, Lin, Zhou, Ran, Xiao, Wu, and Schmidhuber]{hong2024metagptmetaprogrammingmultiagent}
Sirui Hong, Mingchen Zhuge, Jiaqi Chen, Xiawu Zheng, Yuheng Cheng, Ceyao Zhang, Jinlin Wang, Zili Wang, Steven Ka~Shing Yau, Zijuan Lin, Liyang Zhou, Chenyu Ran, Lingfeng Xiao, Chenglin Wu, and Jürgen Schmidhuber.
\newblock Metagpt: Meta programming for a multi-agent collaborative framework, 2024.

\bibitem[Honthaner(2013)]{honthaner2013completeFilmProduction}
Eve~Light Honthaner.
\newblock \emph{The complete film production handbook}.
\newblock Routledge, 2013.

\bibitem[Huang et~al.(2023)Huang, Sun, Xie, Li, and Liu]{huang2023t2i-compbench}
Kaiyi Huang, Kaiyue Sun, Enze Xie, Zhenguo Li, and Xihui Liu.
\newblock T2i-compbench: A comprehensive benchmark for open-world compositional text-to-image generation.
\newblock \emph{Advances in Neural Information Processing Systems}, 36:\penalty0 78723--78747, 2023.

\bibitem[Huang et~al.(2024)Huang, Huang, Ning, Lin, Wang, and Liu]{huang2024genmac}
Kaiyi Huang, Yukun Huang, Xuefei Ning, Zinan Lin, Yu Wang, and Xihui Liu.
\newblock Genmac: Compositional text-to-video generation with multi-agent collaboration.
\newblock \emph{arXiv preprint arXiv:2412.04440}, 2024.

\bibitem[Khachatryan et~al.(2023)Khachatryan, Movsisyan, Tadevosyan, Henschel, Wang, Navasardyan, and Shi]{khachatryan2023text2video}
Levon Khachatryan, Andranik Movsisyan, Vahram Tadevosyan, Roberto Henschel, Zhangyang Wang, Shant Navasardyan, and Humphrey Shi.
\newblock Text2video-zero: Text-to-image diffusion models are zero-shot video generators.
\newblock In \emph{Proceedings of the IEEE/CVF International Conference on Computer Vision}, pages 15954--15964, 2023.

\bibitem[{Kling}(2025)]{kling}
{Kling}.
\newblock \url{https://kling.kuaishou.com/}, 2025.
\newblock Organization: Kuaishou.

\bibitem[Kondratyuk et~al.(2023)Kondratyuk, Yu, Gu, Lezama, Huang, Hornung, Adam, Akbari, Alon, Birodkar, et~al.]{kondratyuk2023videopoet}
Dan Kondratyuk, Lijun Yu, Xiuye Gu, Jos{\'e} Lezama, Jonathan Huang, Rachel Hornung, Hartwig Adam, Hassan Akbari, Yair Alon, Vighnesh Birodkar, et~al.
\newblock Videopoet: A large language model for zero-shot video generation.
\newblock \emph{arXiv preprint arXiv:2312.14125}, 2023.

\bibitem[Kong et~al.(2024)Kong, Tian, Zhang, Min, Dai, Zhou, Xiong, Li, Wu, Zhang, Wu, Lin, Wang, Wang, Li, Huang, Yang, Tan, Wang, Song, Bai, Wu, Xue, Wang, Yuan, Wang, Liu, Li, Li, Wang, Yu, Deng, Li, Long, Chen, Cui, Peng, Yu, He, Xu, Zhou, Xu, Tao, Lu, Liu, Zhou, Wang, Yang, Wang, Liu, Jiang, and Zhong]{kong2024hunyuanvideo}
Weijie Kong, Qi Tian, Zijian Zhang, Rox Min, Zuozhuo Dai, Jin Zhou, Jiangfeng Xiong, Xin Li, Bo Wu, Jianwei Zhang, Kathrina Wu, Qin Lin, Aladdin Wang, Andong Wang, Changlin Li, Duojun Huang, Fang Yang, Hao Tan, Hongmei Wang, Jacob Song, Jiawang Bai, Jianbing Wu, Jinbao Xue, Joey Wang, Junkun Yuan, Kai Wang, Mengyang Liu, Pengyu Li, Shuai Li, Weiyan Wang, Wenqing Yu, Xinchi Deng, Yang Li, Yanxin Long, Yi Chen, Yutao Cui, Yuanbo Peng, Zhentao Yu, Zhiyu He, Zhiyong Xu, Zixiang Zhou, Zunnan Xu, Yangyu Tao, Qinglin Lu, Songtao Liu, Dax Zhou, Hongfa Wang, Yong Yang, Di Wang, Yuhong Liu, Jie Jiang, and Caesar Zhong.
\newblock Hunyuanvideo: A systematic framework for large video generative models, 2024.

\bibitem[Li et~al.(2024)Li, Shi, Hu, Wang, Zhu, Xu, Zhao, and Zhang]{li2024animdirector}
Yunxin Li, Haoyuan Shi, Baotian Hu, Longyue Wang, Jiashun Zhu, Jinyi Xu, Zhen Zhao, and Min Zhang.
\newblock Anim-director: A large multimodal model powered agent for controllable animation video generation.
\newblock In \emph{SIGGRAPH Asia 2024 Conference Papers}, pages 1--11, 2024.

\bibitem[Lin et~al.(2024)Lin, Zala, Cho, and Bansal]{Lin2023VideoDirectorGPT}
Han Lin, Abhay Zala, Jaemin Cho, and Mohit Bansal.
\newblock Videodirectorgpt: Consistent multi-scene video generation via llm-guided planning.
\newblock In \emph{COLM}, 2024.

\bibitem[{LTX Studio}(2024)]{LTXStudioAppMisc}
{LTX Studio}.
\newblock {LTX Studio}.
\newblock \url{https://app.ltx.studio/}, 2024.
\newblock Accessed: 2025-04. Organization: Lightricks.

\bibitem[Luo et~al.(2024)Luo, Zheng, Yang, Li, Lin, Huang, Ji, Chao, Luo, and Ji]{luo2024videoRAG}
Yongdong Luo, Xiawu Zheng, Xiao Yang, Guilin Li, Haojia Lin, Jinfa Huang, Jiayi Ji, Fei Chao, Jiebo Luo, and Rongrong Ji.
\newblock Video-rag: Visually-aligned retrieval-augmented long video comprehension.
\newblock \emph{arXiv preprint arXiv:2411.13093}, 2024.

\bibitem[Luo et~al.(2023)Luo, Chen, Zhang, Huang, Wang, Shen, Zhao, Zhou, and Tan]{luo2023videofusion}
Zhengxiong Luo, Dayou Chen, Yingya Zhang, Yan Huang, Liang Wang, Yujun Shen, Deli Zhao, Jingren Zhou, and Tieniu Tan.
\newblock Videofusion: Decomposed diffusion models for high-quality video generation.
\newblock In \emph{Proceedings of the IEEE/CVF Conference on Computer Vision and Pattern Recognition}, pages 10209--10218, 2023.

\bibitem[Mascelli(1965)]{mascelli1965fiveCofCinematography}
Joseph~V Mascelli.
\newblock \emph{The five C's of cinematography}.
\newblock Grafic Publications Hollywood, 1965.

\bibitem[Murch(2001)]{murch2001blink}
Walter Murch.
\newblock \emph{In the Blink of an Eye}.
\newblock Silman-James Press Los Angeles, 2001.

\bibitem[OpenAI(2024)]{gpt-4o}
OpenAI.
\newblock {Hello GPT-4o}.
\newblock \url{https://openai.com/index/hello-gpt-4o/}, 2024.
\newblock Accessed: 2024-11-14.

\bibitem[OpenTimelineIO(2023)]{openTimelineIO}
OpenTimelineIO.
\newblock Opentimelineio: An open-source framework for interchange of editorial timeline data.
\newblock \url{https://opentimelineio.readthedocs.io/}, 2023.
\newblock Accessed: 2025-05-09.

\bibitem[Park et~al.(2021)Park, Azadi, Liu, Darrell, and Rohrbach]{park2021benchmark}
Dong~Huk Park, Samaneh Azadi, Xihui Liu, Trevor Darrell, and Anna Rohrbach.
\newblock Benchmark for compositional text-to-image synthesis.
\newblock In \emph{NeurIPS}, 2021.

\bibitem[Park et~al.(2023)Park, O'Brien, Cai, Morris, Liang, and Bernstein]{park2023generativeagentsinteractivesimulacra}
Joon~Sung Park, Joseph~C. O'Brien, Carrie~J. Cai, Meredith~Ringel Morris, Percy Liang, and Michael~S. Bernstein.
\newblock Generative agents: Interactive simulacra of human behavior, 2023.

\bibitem[Rabiger(2013)]{rabiger2013directing}
Michael Rabiger.
\newblock \emph{Directing: Film techniques and aesthetics}.
\newblock Routledge, 2013.

\bibitem[Ruszev(2018)]{ruszev2018rhythmic}
Szilvia Ruszev.
\newblock Rhythmic trajectories--visualizing cinematic rhythm in film sequences.
\newblock \emph{Women Cutting Movies: Editors from East and Central Europe. Special Issue of Apparatus. Film Media and Digital Cultures of Central and Eastern Europe}, 7, 2018.

\bibitem[Singer et~al.(2022)Singer, Polyak, Hayes, Yin, An, Zhang, Hu, Yang, Ashual, Gafni, et~al.]{singer2022make}
Uriel Singer, Adam Polyak, Thomas Hayes, Xi Yin, Jie An, Songyang Zhang, Qiyuan Hu, Harry Yang, Oron Ashual, Oran Gafni, et~al.
\newblock Make-a-video: Text-to-video generation without text-video data.
\newblock \emph{arXiv preprint arXiv:2209.14792}, 2022.

\bibitem[Sun et~al.(2024)Sun, Yin, Li, Wu, Qiu, and Kong]{sun2024corexpushingboundariescomplex}
Qiushi Sun, Zhangyue Yin, Xiang Li, Zhiyong Wu, Xipeng Qiu, and Lingpeng Kong.
\newblock Corex: Pushing the boundaries of complex reasoning through multi-model collaboration, 2024.

\bibitem[Team et~al.(2023)Team, Anil, Borgeaud, Alayrac, Yu, Soricut, Schalkwyk, Dai, Hauth, Millican, et~al.]{team2023gemini}
Gemini Team, Rohan Anil, Sebastian Borgeaud, Jean-Baptiste Alayrac, Jiahui Yu, Radu Soricut, Johan Schalkwyk, Andrew~M Dai, Anja Hauth, Katie Millican, et~al.
\newblock Gemini: a family of highly capable multimodal models.
\newblock \emph{arXiv preprint arXiv:2312.11805}, 2023.

\bibitem[Villegas et~al.(2022)Villegas, Babaeizadeh, Kindermans, Moraldo, Zhang, Saffar, Castro, Kunze, and Erhan]{villegas2022phenaki}
Ruben Villegas, Mohammad Babaeizadeh, Pieter-Jan Kindermans, Hernan Moraldo, Han Zhang, Mohammad~Taghi Saffar, Santiago Castro, Julius Kunze, and Dumitru Erhan.
\newblock Phenaki: Variable length video generation from open domain textual descriptions.
\newblock In \emph{International Conference on Learning Representations}, 2022.

\bibitem[Wang et~al.(2025)Wang, Ai, Wen, Mao, Xie, Chen, Yu, Zhao, Yang, Zeng, Wang, Zhang, Zhou, Wang, Chen, Zhu, Zhao, Yan, Huang, Feng, Zhang, Li, Wu, Chu, Feng, Zhang, Sun, Fang, Wang, Gui, Weng, Shen, Lin, Wang, Wang, Zhou, Wang, Shen, Yu, Shi, Huang, Xu, Kou, Lv, Li, Liu, Wang, Zhang, Huang, Li, Wu, Liu, Pan, Zheng, Hong, Shi, Feng, Jiang, Han, Wu, and Liu]{wan2025}
Ang Wang, Baole Ai, Bin Wen, Chaojie Mao, Chen-Wei Xie, Di Chen, Feiwu Yu, Haiming Zhao, Jianxiao Yang, Jianyuan Zeng, Jiayu Wang, Jingfeng Zhang, Jingren Zhou, Jinkai Wang, Jixuan Chen, Kai Zhu, Kang Zhao, Keyu Yan, Lianghua Huang, Mengyang Feng, Ningyi Zhang, Pandeng Li, Pingyu Wu, Ruihang Chu, Ruili Feng, Shiwei Zhang, Siyang Sun, Tao Fang, Tianxing Wang, Tianyi Gui, Tingyu Weng, Tong Shen, Wei Lin, Wei Wang, Wei Wang, Wenmeng Zhou, Wente Wang, Wenting Shen, Wenyuan Yu, Xianzhong Shi, Xiaoming Huang, Xin Xu, Yan Kou, Yangyu Lv, Yifei Li, Yijing Liu, Yiming Wang, Yingya Zhang, Yitong Huang, Yong Li, You Wu, Yu Liu, Yulin Pan, Yun Zheng, Yuntao Hong, Yupeng Shi, Yutong Feng, Zeyinzi Jiang, Zhen Han, Zhi-Fan Wu, and Ziyu Liu.
\newblock Wan: Open and advanced large-scale video generative models.
\newblock \emph{arXiv preprint arXiv:2503.20314}, 2025.

\bibitem[Wang et~al.(2023)Wang, Yuan, Chen, Zhang, Wang, and Zhang]{wang2023modelscope}
Jiuniu Wang, Hangjie Yuan, Dayou Chen, Yingya Zhang, Xiang Wang, and Shiwei Zhang.
\newblock Modelscope text-to-video technical report.
\newblock \emph{arXiv preprint arXiv:2308.06571}, 2023.

\bibitem[Wei et~al.(2022)Wei, Tay, Bommasani, Raffel, Zoph, Borgeaud, Yogatama, Bosma, Zhou, Metzler, et~al.]{wei2022emergent}
Jason Wei, Yi Tay, Rishi Bommasani, Colin Raffel, Barret Zoph, Sebastian Borgeaud, Dani Yogatama, Maarten Bosma, Denny Zhou, Donald Metzler, et~al.
\newblock Emergent abilities of large language models.
\newblock \emph{arXiv preprint arXiv:2206.07682}, 2022.

\bibitem[Wu et~al.(2023)Wu, Bansal, Zhang, Wu, Li, Zhu, Jiang, Zhang, Zhang, Liu, Awadallah, White, Burger, and Wang]{wu2023autogenenablingnextgenllm}
Qingyun Wu, Gagan Bansal, Jieyu Zhang, Yiran Wu, Beibin Li, Erkang Zhu, Li Jiang, Xiaoyun Zhang, Shaokun Zhang, Jiale Liu, Ahmed~Hassan Awadallah, Ryen~W White, Doug Burger, and Chi Wang.
\newblock Autogen: Enabling next-gen llm applications via multi-agent conversation, 2023.

\bibitem[Wu et~al.(2025)Wu, Zhu, and Shou]{wu2025movieagent}
Weijia Wu, Zeyu Zhu, and Mike~Zheng Shou.
\newblock Automated movie generation via multi-agent cot planning.
\newblock \emph{arXiv preprint arXiv:2503.07314}, 2025.

\bibitem[Xu et~al.(2025)Xu, Wang, Wang, Li, Shi, Yang, Wang, Hu, Yu, and Zhang]{xu2025filmagent}
Zhenran Xu, Longyue Wang, Jifang Wang, Zhouyi Li, Senbao Shi, Xue Yang, Yiyu Wang, Baotian Hu, Jun Yu, and Min Zhang.
\newblock Filmagent: A multi-agent framework for end-to-end film automation in virtual 3d spaces.
\newblock \emph{arXiv preprint arXiv:2501.12909}, 2025.

\bibitem[Yang et~al.(2024)Yang, Teng, Zheng, Ding, Huang, Xu, Yang, Hong, Zhang, Feng, et~al.]{yang2024cogvideox}
Zhuoyi Yang, Jiayan Teng, Wendi Zheng, Ming Ding, Shiyu Huang, Jiazheng Xu, Yuanming Yang, Wenyi Hong, Xiaohan Zhang, Guanyu Feng, et~al.
\newblock Cogvideox: Text-to-video diffusion models with an expert transformer.
\newblock \emph{arXiv preprint arXiv:2408.06072}, 2024.

\bibitem[Yu et~al.(2023{\natexlab{a}})Yu, Cheng, Sohn, Lezama, Zhang, Chang, Hauptmann, Yang, Hao, Essa, et~al.]{yu2023magvit}
Lijun Yu, Yong Cheng, Kihyuk Sohn, Jos{\'e} Lezama, Han Zhang, Huiwen Chang, Alexander~G Hauptmann, Ming-Hsuan Yang, Yuan Hao, Irfan Essa, et~al.
\newblock Magvit: Masked generative video transformer.
\newblock In \emph{Proceedings of the IEEE/CVF Conference on Computer Vision and Pattern Recognition}, pages 10459--10469, 2023{\natexlab{a}}.

\bibitem[Yu et~al.(2023{\natexlab{b}})Yu, Lezama, Gundavarapu, Versari, Sohn, Minnen, Cheng, Gupta, Gu, Hauptmann, et~al.]{yu2023language}
Lijun Yu, Jos{\'e} Lezama, Nitesh~B Gundavarapu, Luca Versari, Kihyuk Sohn, David Minnen, Yong Cheng, Agrim Gupta, Xiuye Gu, Alexander~G Hauptmann, et~al.
\newblock Language model beats diffusion--tokenizer is key to visual generation.
\newblock \emph{arXiv preprint arXiv:2310.05737}, 2023{\natexlab{b}}.

\bibitem[Yuan et~al.(2024)Yuan, Liu, Cao, Sun, Jia, Chen, Li, Lin, Yuan, He, Wang, Ye, and Sun]{yuan2024moraenablinggeneralistvideo}
Zhengqing Yuan, Yixin Liu, Yihan Cao, Weixiang Sun, Haolong Jia, Ruoxi Chen, Zhaoxu Li, Bin Lin, Li Yuan, Lifang He, Chi Wang, Yanfang Ye, and Lichao Sun.
\newblock Mora: Enabling generalist video generation via a multi-agent framework, 2024.

\bibitem[Zhang et~al.(2025)Zhang, Yu, Min, Xin, Wei, Shi, Huang, Kong, Xin, Jiang, et~al.]{zhang2025generativeAIforFilm}
Ruihan Zhang, Borou Yu, Jiajian Min, Yetong Xin, Zheng Wei, Juncheng~Nemo Shi, Mingzhen Huang, Xianghao Kong, Nix~Liu Xin, Shanshan Jiang, et~al.
\newblock Generative ai for film creation: A survey of recent advances.
\newblock \emph{arXiv preprint arXiv:2504.08296}, 2025.

\bibitem[Zheng et~al.(2025)Zheng, Huang, Liu, Zou, He, Zhang, Zhang, He, Zheng, Qiao, et~al.]{zheng2025vbench2}
Dian Zheng, Ziqi Huang, Hongbo Liu, Kai Zou, Yinan He, Fan Zhang, Yuanhan Zhang, Jingwen He, Wei-Shi Zheng, Yu Qiao, et~al.
\newblock Vbench-2.0: Advancing video generation benchmark suite for intrinsic faithfulness.
\newblock \emph{arXiv preprint arXiv:2503.21755}, 2025.

\bibitem[Zhou et~al.(2022)Zhou, Wang, Yan, Lv, Zhu, and Feng]{zhou2022magicvideo}
Daquan Zhou, Weimin Wang, Hanshu Yan, Weiwei Lv, Yizhe Zhu, and Jiashi Feng.
\newblock Magicvideo: Efficient video generation with latent diffusion models.
\newblock \emph{arXiv preprint arXiv:2211.11018}, 2022.

\bibitem[Zhuang et~al.(2025)Zhuang, Huang, Cheng, Wu, Hu, Liao, Huang, Wang, Liao, Cai, Xu, Zhang, Zeng, Yu, and Zhang]{zhuang2025vistorybench}
Cailin Zhuang, Ailin Huang, Wei Cheng, Jingwei Wu, Yaoqi Hu, Jiaqi Liao, Zhewei Huang, Hongyuan Wang, Xinyao Liao, Weiwei Cai, Hengyuan Xu, Xuanyang Zhang, Xianfang Zeng, Gang Yu, and Chi Zhang.
\newblock {ViStoryBench: A Comprehensive Benchmark Suite for Story Visualization}.
\newblock \emph{arXiv preprint arXiv:2505.24862}, 2025.

\end{thebibliography}
}

% Appendix
% \appendix

% \clearpage
% \input{sec/6_fig}

\clearpage
\appendix

\section{Scene Structuring}
\label{app: method_scene_structuring}
Given an input theme sentence, the script is hierarchically expanded.
% —evolving from high-level narrative structure to detailed visual elements. 
This hierarchical approach ensures coherent progression from abstract concepts to concrete visual descriptions that can be translated into audiovisual content. 
% The overall goal is to structure scenes based on coherent time-space blocks, which serve as basic audiovisual design units for following steps and can include multiple shots, aligning with professional film production workflows.
The scene structuring unfolds in four progressive phases, each building upon the output of the previous phase, implemented through a chain of four specialized LLMs working sequentially:
\begin{itemize}
    \item \textbf{Synopsis:} The input theme is expanded into a synopsis that introduces settings, characters, and plots, forming the foundation for subsequent development. This phase establishes the core narrative framework without detailed visual considerations. 
    \item \textbf{Simplified Storyboard:} The synopsis is enriched with background information, causal chains, narrative developments, and outcomes, ensuring the story has appropriate twists while maintaining narrative coherence and logical consistency. This phase transforms narrative concepts into structured story events.
    \item \textbf{Detailed Storyboard:} The simplified version is further elaborated into a detailed storyboard containing concrete visual narratives. This includes both primary shots (featuring main character actions with specific descriptions) and contextual shots (showing surrounding environments and establishing shots), alongside metaphorical imagery to enhance emotional resonance and thematic depth. This phase translates story events into visual sequences.
    \item \textbf{Scene Block:} The detailed storyboard is segmented into distinct scene blocks that function as spatio-temporal contexts within one scene (\textit{i.e.}, events occurring in the same location during the same timeframe), based on chronological order and spatial continuity. These blocks serve as fundamental units for subsequent audiovisual generation, ensuring consistent visual and narrative treatment across related shots. For each block, we extract key elements including prompt, time, location, characters, visual elements, and narrative objectives (justifying each scene's narrative necessity for plot advancement or character development). These extracted elements collectively provide structured input for the Scene-level Multi-shot Camera Design module.
    Besides, we use LLM to add rough sound design by text description for each scene block, used for audience-centric review and scene-level sound design (background ambiance and music scoring) in \coordination{}.
\end{itemize}

\section{Sound Design}
\label{app: sound_design}
Directly relying on MLLMs to design multi-track audio (including background ambience, musical scoring, voice-over (VO), foley, and sound effects) often results in poor synchronization with video content. This is primarily due to the LLM's limited capability in managing coordination across multiple audio layers and temporal alignment. 
To address this, we propose a multi-scale audio-visual synchronization strategy, which systematically designs different audio types at appropriate temporal resolutions. This process is further supported by retrieval-augmented generation (RAG) from curated audio libraries or synthesized voice content, followed by systematic audio preprocessing.

Specifically, synchronization is designed across three temporal scales: 
\begin{itemize} 
\item Scene-level synchronization: Background ambience and musical scoring are assigned at the scene level to establish atmosphere and emotional tone. This builds upon the rough sound design from Scene Structuring. 
\item Shot-level synchronization: VO, including narration and dialogue, is designed at the shot level to maintain close alignment with visual content and enhance narrative clarity. 
\item Intra-shot synchronization: Foley and sound effects are assigned at a fine-grained temporal resolution. MLLM analyzes video content with second-level timecodes to detect specific actions, movements, and environmental elements that require sonic representation, ensuring detailed audiovisual coordination. 

\end{itemize}

To ensure audio quality and contextual relevance, we employ the RAG approach for non-voice audio tracks. We construct a curated audio library comprising 46,826 sound assets, including 5,877 music tracks and 40,949 other audio assets (atmospheric sounds, foley, and effects) sourced from the Internet. 
Each asset is tagged with semantic descriptors, emotional qualities, and acoustic properties to facilitate context-aware retrieval.
For voice-over content, we utilize ElevenLabs' text-to-speech services, allowing for flexible and high-fidelity voice generation tailored to specific narrative demands.

Composing multiple audio tracks directly often leads to inconsistencies in loudness, frequency balance, and dynamic range. To mitigate these issues, we apply audio preprocessing techniques that ensure cross-track and track-video harmonization. This includes: 
\begin{itemize} 
\item Loudness units relative to full scale (LUFS) normalization to maintain consistent perceived loudness across tracks (\textit{e.g.} -16 LUFS for voice content, -28 LUFS for background elements). 
\item Frequency adjustment to avoid spectral clashes between different types of audio, particularly ensuring voice intelligibility by attenuating competing frequencies in background tracks. 
\item Dynamic equalization to enhance clarity and cohesion. 
\end{itemize}

% \section{Evaluation Metrics}
% \label{app:evaluation_explanation}

\section{\eval{} Evaluation Instructions}
\label{app: evaluation_instructions}
We show the detailed criteria of \eval{} for both automatic evaluation and user study as follows: 

\clearpage

% --- Assume your 'example' environment is defined in the preamble ---
% --- and that acmart is okay with its usage and internal paragraph spacing ---

% ==================================================
% NARRATIVE & SCRIPT
% ==================================================
\begin{example}[{\small NARRATIVE AND SCRIPT}]
% If you want the content of the example also to be small, uncomment the next line
% \small

\noindent\textbf{Script Faithfulness}
\par % Start new paragraph for the list of points
\textbf{1 point:} Severely deviates from the original script; scenes and character settings completely inconsistent with the original.
\par
\textbf{2 points:} Partially follows the original script, but with obvious deviations; multiple key settings changed.
\par
\textbf{3 points:} Generally follows the original script, preserving main scenes and settings, but with details omitted.
\par
\textbf{4 points:} Highly faithful to the original script, accurately reproduces most scenes and settings with rich details.
\par
\textbf{5 points:} Completely faithful to the original script, precisely presents all scenes, settings, and details.

\par\medskip % Space before next sub-category

\noindent\textbf{Narrative Coherence}
\par
\textbf{1 point:} Chaotic and disorderly story with serious logical contradictions and plot discontinuities.
\par
\textbf{2 points:} Story is basically understandable but contains multiple obvious logical gaps and coherence issues.
\par
\textbf{3 points:} Story is generally coherent with minor logical deficiencies that don't affect understanding of the main plot.
\par
\textbf{4 points:} Story flows smoothly and coherently with reasonable plot development and almost no obvious logical issues.
\par
\textbf{5 points:} Story is completely coherent with natural and reasonable plot development, clear cause-effect relationships, and no logical holes.
\end{example}

\FloatBarrier 

% ==================================================
% AUDIO-VISUAL & TECHNICAL
% ==================================================
\begin{example}[{\small AUDIO-VISUALS AND TECHNIQUES}]
% \small % Uncomment if content should be small

\noindent\textbf{Visual Quality}
\par
\textbf{1 point:} Severely broken visuals with numerous missing or distorted visual elements.
\par
\textbf{2 points:} Obvious visual flaws with some scenes showing missing or distorted elements.
\par
\textbf{3 points:} Basically complete visuals with occasional minor errors that don't affect overall viewing.
\par
\textbf{4 points:} Clear and complete visuals with very few minor imperfections.
\par
\textbf{5 points:} Flawless visuals with all elements perfectly rendered and no visual breakdowns.

\par\medskip % Space before next sub-category

\noindent\textbf{Character Consistency (visual)}
\par
\textbf{1 point:} Severely inconsistent character designs with dramatic appearance changes of the same character across different scenes.
\par
\textbf{2 points:} Noticeable fluctuations in character designs with character features changing in some scenes.
\par
\textbf{3 points:} Generally consistent character designs with occasional minor inconsistencies that aren't obvious.
\par
\textbf{4 points:} Highly consistent character designs that maintain stable features across different scenes and angles.
\par
\textbf{5 points:} Perfectly consistent character designs that maintain precise character features in all scenes and actions.

\par\medskip % Space before next sub-category

\noindent\textbf{Physical Law Compliance}
\par
\textbf{1 point:} Severely violates physical laws with extremely unnatural movements, collisions, and effects.
\par
\textbf{2 points:} Multiple violations of basic physical laws with obviously unrealistic movements and effects.
\par
\textbf{3 points:} Generally complies with physical laws; some movements or effects appear slightly artificial but acceptable.
\par
\textbf{4 points:} Good compliance with physical laws, natural movements, and believable effects.
\par
\textbf{5 points:} Perfect compliance with physical laws where all movements, collisions, and effects are extremely realistic.

\par\medskip % Space before next sub-category

\noindent\textbf{Voice/Audio Quality}
\par
\textbf{1 point:} Extremely poor audio quality with unclear voiceovers and chaotic or missing sound effects.
\par
\textbf{2 points:} Poor audio quality with partially unclear voiceovers and simple or inadequate sound effects.
\par
\textbf{3 points:} Average audio quality with basically clear voiceovers and appropriate but not outstanding sound effects.
\par
\textbf{4 points:} Good audio quality with clear voiceovers and rich sound effects that match the scenes.
\par
\textbf{5 points:} Excellent audio quality with extremely clear and vivid voiceovers, and rich, nuanced sound effects with great expressiveness.
\end{example}

\clearpage
% ==================================================
% AESTHETICS & EXPRESSION
% ==================================================
\begin{example}[{\small AESTHETICS AND EXPRESSION}]
% \small % Uncomment if content should be small

\noindent\textbf{Cinematic Techniques}
\par
\textbf{1 point:} Single, stiff shots with no variation and lack of basic film language.
\par
\textbf{2 points:} Limited shot variation, stiff camera movements, and poor film language.
\par
\textbf{3 points:} Uses common shot techniques with basically smooth camera movements and basic film language expression.
\par
\textbf{4 points:} Rich film language, smooth and natural camera movements, reasonable and effective shot variations.
\par
\textbf{5 points:} Highly creative shot usage, precise camera movements, and rich shot variations with exceptional expressiveness.

\par\medskip % Space before next sub-category

\noindent\textbf{Audio-Visual Richness}
\par
\textbf{1 point:} Extremely limited visual and auditory expression. The film uses monotonous or repetitive visual/audio elements with minimal variation or artistic layering.
\par
\textbf{2 points:} Some attempts at audiovisual expression, but overall lacks creativity. Techniques are basic and formulaic, offering little dynamic or stylistic variation.
\par
\textbf{3 points:} Moderate diversity in audiovisual methods. Certain scenes explore varied styles or tempos, but the richness is uneven and lacks coherence or artistic depth.
\par
\textbf{4 points:} Visually and sonically expressive. Multiple techniques are used effectively to create layered meaning, mood shifts, or stylistic nuances across the film.
\par
\textbf{5 points:} Exceptionally rich audiovisual language. Diverse, inventive, and highly expressive use of both sound and visuals creates a compelling and distinctive artistic voice with strong narrative or emotional impact.
\end{example}

% ==================================================
% AUDIENCE EXPERIENCE DIMENSIONS
% ==================================================
\begin{example}[{\small RHYTHM AND FLOW}]
% \small % Uncomment if content should be small

\noindent\textbf{Narrative Pacing}
\par
\textbf{1 point:} Completely uncontrolled pacing, either too fast or too slow, severely affecting story comprehension.
\par
\textbf{2 points:} Obviously inconsistent pacing with some plot developments that are too quick or too slow.
\par
\textbf{3 points:} Generally appropriate pacing with reasonable progression of main plot elements.
\par
\textbf{4 points:} Well-controlled pacing with natural plot progression and good balance between tension and relief.
\par
\textbf{5 points:} Precisely controlled pacing that both serves the story needs and captures audience emotions, with perfect balance.

\par\medskip % Space before next sub-category

\noindent\textbf{Video-Audio Coordination}
\par
\textbf{1 point:} Severely unsynchronized audio and video with completely mismatched lip-syncing.
\par
\textbf{2 points:} Obvious lack of synchronization between audio and video with poor coordination between voice and visuals.
\par
\textbf{3 points:} Basically synchronized audio and video with occasional inconsistencies that don't obviously interfere with viewing.
\par
\textbf{4 points:} Good audio-video coordination with highly matched sound and visual actions.
\par
\textbf{5 points:} Perfect synchronization where all sound elements precisely match visual actions, creating a harmonious viewing experience.
\end{example}

% ==================================================
% EMOTIONAL & ENGAGEMENT
% ==================================================
\begin{example}[{\small EMOTIONAL AND ENGAGEMENT}]
% \small % Uncomment if content should be small

\noindent\textbf{Compelling Degree}
\par
\textbf{1 point:} No appeal whatsoever; difficult for viewers to feel immersed or emotionally connected.
\par
\textbf{2 points:} Insufficient appeal with weak emotional rendering; difficult to maintain viewer attention.
\par
\textbf{3 points:} Basic appeal that can generate viewer interest but insufficient to create profound emotional resonance.
\par
\textbf{4 points:} Strong appeal with effective emotional rendering that can evoke obvious emotional resonance from viewers.
\par
\textbf{5 points:} Extremely compelling with powerful emotional tension that fully engages viewers throughout and creates strong emotional resonance.
\end{example}

% ==================================================
% OVERALL EXPERIENCE
% ==================================================
\clearpage
\begin{example}[{\small OVERALL EXPERIENCE}]
% \small % Uncomment if content should be small

\noindent\textbf{Overall Quality}
\par
\textbf{1 point:} Extremely poor quality with multiple dimensions severely below standard; completely lacks viewing value.
\par
\textbf{2 points:} Poor quality with main dimensions performing badly; limited viewing value.
\par
\textbf{3 points:} Average quality with average performance across dimensions; has basic viewing value.
\par
\textbf{4 points:} Good quality with good performance across dimensions that work well together; has high viewing value.
\par
\textbf{5 points:} Excellent quality with outstanding performance across all dimensions and perfect coordination; has extremely high artistic and viewing value.
\end{example}

\section{Qualitative Results}
\begin{figure*}[htbp] % Using htbp for more placement flexibility
  \centering
  
  % First image
  \includegraphics[width=\linewidth]{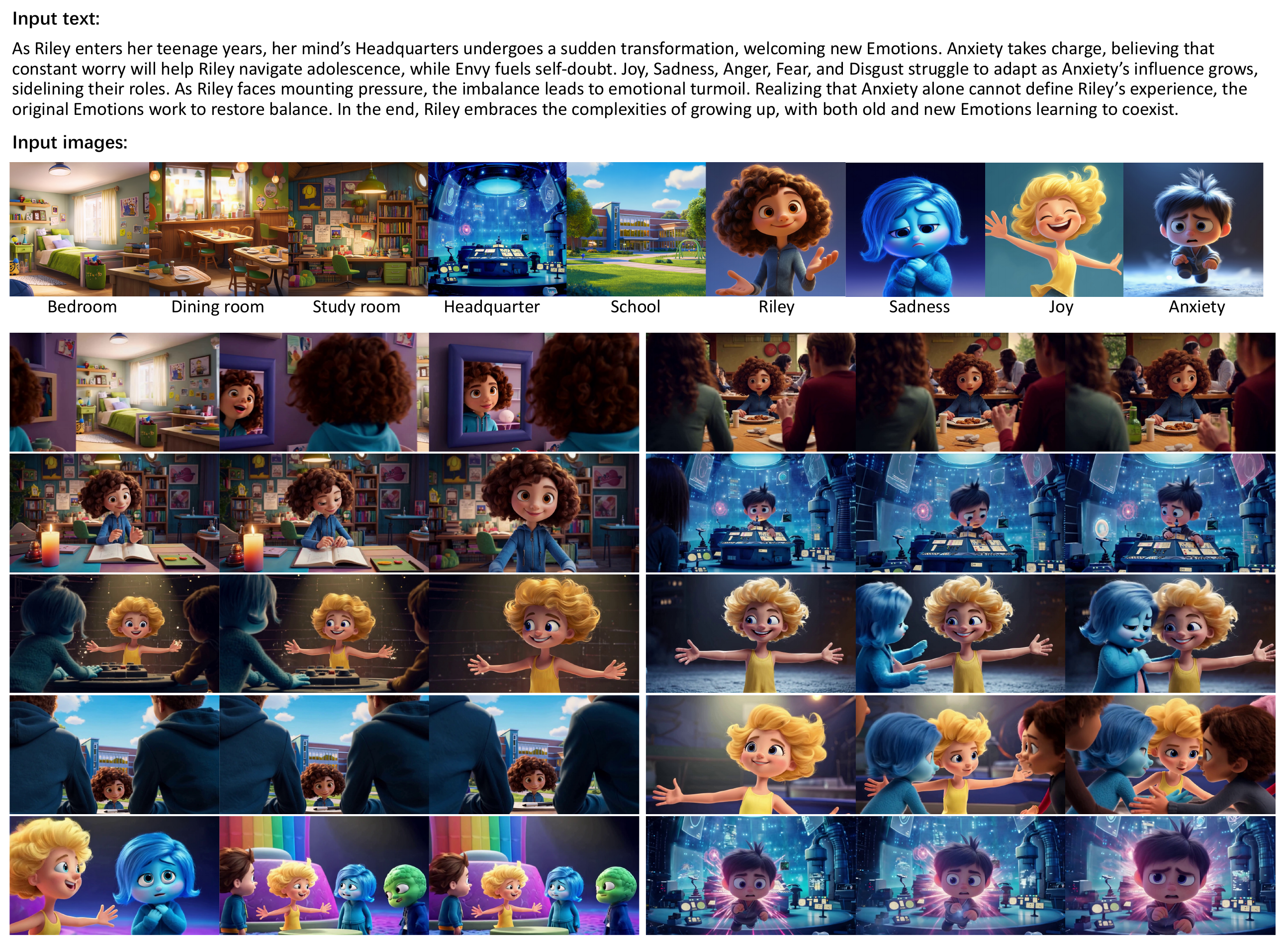}
  % \vspace{1ex} % Optional: add a little vertical space between images if needed
  
  % Second image
  \includegraphics[width=\linewidth]{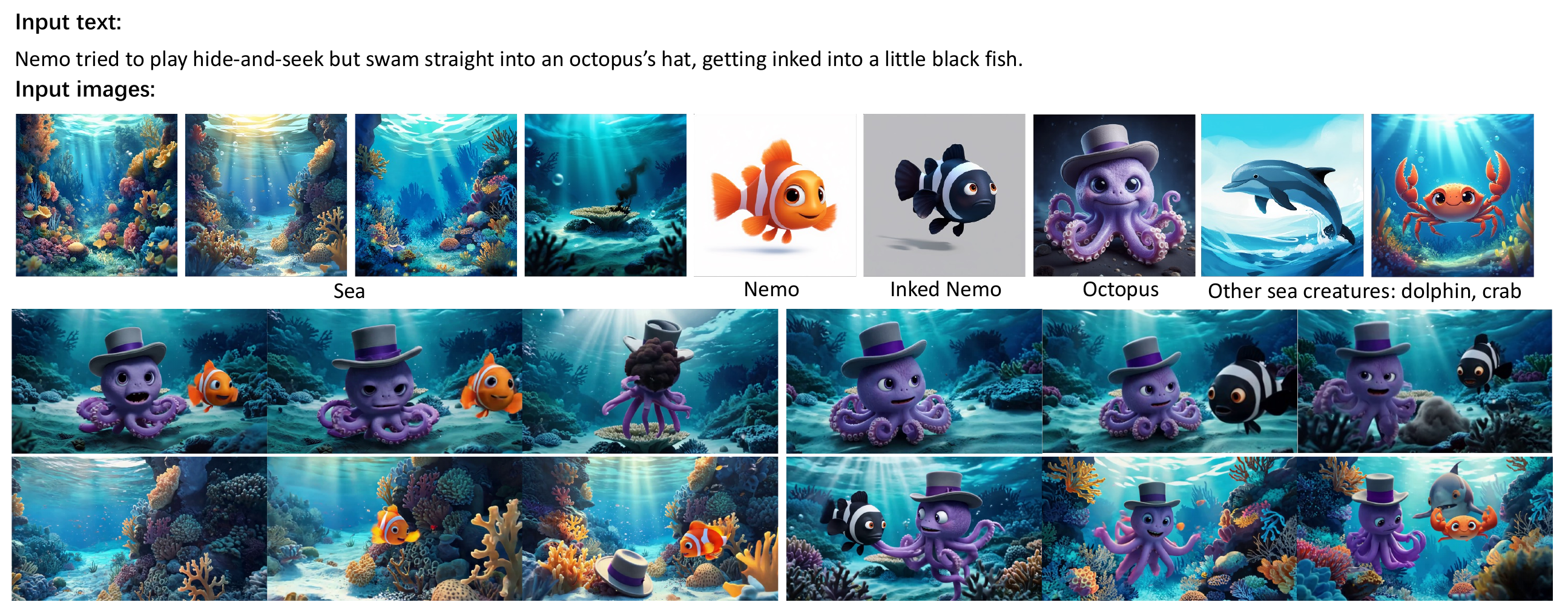}
  
  % \vspace{-10pt} % Optional: adjust space before caption if needed
  \caption{
  Qualitative results of our \name{}. Our \name{} can generate high-quality films with either simple or complex script input.
  }

  \label{fig:combined_results} % Use a new, unique label for the combined figure
  % \vspace{-15pt} % Optional: adjust space after figure if needed
\end{figure*}

We show the qualitative results in~\Cref{fig:combined_results}. Our method can generate high-quality films with both simple and complex script input.

% 12个维度的具体标准

% \subsection{Ablation Study}
% detailed score
% 详细说明：
% use one case (little prince)
% baseline是只有detailed prompt，没有time space block
% audience perspective和editing是绑定在一起的
% audio-video sync是只有audio composition step 3的scene-level设计
% \input{tables/ablation_detailed}

\section{Quantitative Results}
% \subsection{Quantitative Results}
We show the detailed results of user study in~\Cref{tab:human_eval_detail}, and detailed human correlation in twelve criteria in \eval{} in ~\Cref{tab:human_corr_detail}. Our method shows superior performances compared with other existing film generation systems. Our proposed automatic evaluation shows high correlation with human judgement over 12 evaluation criteria.
\begin{table*}[h]
\centering
\caption{
Quantitative comparison of different methods across 12 evaluation criteria.
The ``avg'' column shows the average score for each method.
Values in bold indicate the best performance in that column.
}
\label{tab:human_eval_detail} % Changed label slightly
\resizebox{\linewidth}{!}{
\begin{tabular}{l cc cccc cc cc c c c}
\toprule
\multirow{2}{*}{Method} & \multicolumn{2}{c}{NS $\uparrow$} & \multicolumn{4}{c}{AT $\uparrow$} & \multicolumn{2}{c}{AE $\uparrow$} & \multicolumn{2}{c}{RF $\uparrow$} & EE $\uparrow$ & OE $\uparrow$ &\multirow{2}{*}{Avg $\uparrow$} \\
\cmidrule(lr){2-3} \cmidrule(lr){4-7} \cmidrule(lr){8-9} \cmidrule(lr){10-11} \cmidrule(lr){12-12} \cmidrule(lr){13-13}  
 & SF & NC & VQ & CC & PLC & V/AQ & CT & AVR & NP & VAC & CD & OQ &  \\

% \textbf{Ours\textsuperscript{*}} & \textbf{4.00} & \textbf{3.80} & \textbf{4.00} & \textbf{4.40} & \textbf{3.80} & \textbf{3.80} & \textbf{4.00} & \textbf{4.20} & \textbf{4.00} & \textbf{3.80} & \textbf{4.40} & \textbf{4.00} & \textbf{4.02} \\
\midrule
Anim-Director & 2.08 & 1.80 & 2.52 & 2.24 & 2.32 & 1.56 & 2.20 & 1.68 & 2.64 & 1.60 & 2.12 & 2.36 & 2.09 \\
MovieAgent & 1.73 & 1.40 & 1.67 & 1.73 & 1.73 & 1.40 & 1.93 & 1.47 & 2.00 & 1.40 & 2.20 & 2.27 & 1.74 \\
LTX-Studio\textsuperscript{*}  & 2.44 & 2.12 & 3.16 & 3.00 & 2.84 & 3.16 & 3.52 & 2.92 & 2.60 & 3.20 & 3.16 & 2.96 & 2.92 \\
\textbf{Ours} & \textbf{3.73} & \textbf{3.67} & \textbf{3.87} & \textbf{3.93} & \textbf{3.53} & \textbf{3.87} & \textbf{3.73} & \textbf{3.87} & \textbf{3.93} & \textbf{3.53} & \textbf{3.93} & \textbf{3.87} & \textbf{3.79} \\

\bottomrule
\end{tabular}%
}
\begin{flushleft}
\footnotesize
% Any additional notes can go here.
% For example: M1-M12 are shorthand for evaluation metrics.
\end{flushleft}
\end{table*}

\begin{table*}[h]
\centering
\caption{
Quantitative results of correlation coefficients across 12 evaluation points and their average in terms of Pearson Correlation $r$, Spearman's $\rho$, and Kendall's $\tau$ Coefficient ($p$-value < 0.01). \colorbox{mycolor_green}{Green} represents the average of each correlation coefficient across 12 evaluation dimension.
% You can adapt this caption further if "1-12" have specific meanings.
% Example: Quantitative comparison of correlation coefficients for [Your Task/Metric] across 12 [Items/Conditions/etc.] on \eval{DatasetName}.
}
\label{tab:human_corr_detail} 
\resizebox{\linewidth}{!}{
\begin{tabular}{l cc cccc cc cc c c c}
\toprule
% 第一行表头
\multirow{2}{*}{Correlation} & \multicolumn{2}{c}{NS} & \multicolumn{4}{c}{AT } & \multicolumn{2}{c}{AE } & \multicolumn{2}{c}{RF} & EE  & OE  &  \\ % 表头第一行 "Avg"
\cmidrule(lr){2-3} \cmidrule(lr){4-7} \cmidrule(lr){8-9} \cmidrule(lr){10-11} \cmidrule(lr){12-12} \cmidrule(lr){13-13}  
% 第二行表头
 & SF & NC & VQ & CC & PLC & V/AQ & CT & AVR & NP & VAC & CD & OQ & \multicolumn{1}{c}{\cellcolor{mycolor_green}{Avg}}\\ % MODIFIED LINE: 显式为 \multirow 下方的单元格上色
\midrule
$r$ ($\uparrow$) & 0.5818 & 0.6734 & 0.6731 & 0.7114 & 0.4664 & 0.7892 & 0.6301 & 0.5749 & 0.6129 & 0.7964 & 0.5717 & 0.6758 & \cellcolor{mycolor_green}{0.6464} \\
$\rho$ ($\uparrow$) & 0.5842 & 0.6676 & 0.6756 & 0.7184 & 0.4328 & 0.8036 & 0.6565 & 0.5867 & 0.6349 & 0.7815 & 0.5673 & 0.6827 & \cellcolor{mycolor_green}{0.6493} \\
$\tau$ ($\uparrow$) & 0.5056 & 0.5864 & 0.6249 & 0.6474 & 0.4003 & 0.7093 & 0.5662 & 0.4928 & 0.5569 & 0.6973 & 0.4893 & 0.6023 & \cellcolor{mycolor_green}{0.5732} \\
\bottomrule
\end{tabular}
}
% \begin{flushleft} % Optional: if you have notes for this table
% \footnotesize
% Any notes specific to this table can go here.
% \end{flushleft}
\end{table*}

% \subsection{Qualitative Results}
% camera design difference 

\section{Limitations}
While \name{} represents a significant step towards automated professional film generation, it currently has certain limitations. For instance, advanced post-production techniques such as color grading and a diverse range of cinematic transitions are not yet incorporated into our system. These aspects, crucial for achieving a fully polished cinematic look and feel, were beyond the primary scope of this work, which focused on foundational camera language design and cinematic rhythm control. We acknowledge their importance and plan to address their integration in future research.

\end{document}